\documentclass[sigconf,authorversion,nonacm]{acmart}

\AtBeginDocument{%
  \providecommand\BibTeX{{%
    \normalfont B\kern-0.5em{\scshape i\kern-0.25em b}\kern-0.8em\TeX}}}

\providetoggle{editing_mode}
\settoggle{editing_mode}{true} 

\usepackage{tikz,amsmath,xcolor,url,xspace,comment}
\usepackage{subfig}
\usepackage{array}
\usepackage[linesnumbered,ruled,vlined]{algorithm2e}
\usepackage{listings}
\usepackage[normalem]{ulem}
\usepackage{makecell}
\usepackage{multirow, multicol}
\usepackage{hyperref}
\usepackage{soul}
\usepackage{pifont}
\usepackage[skip=3pt,font=small]{caption} 
\usepackage{balance}

\graphicspath{{./fig/}}
\DeclareGraphicsExtensions{.pdf,.png,.eps}
\definecolor{codegreen}{rgb}{0,0.6,0}
\definecolor{codegray}{rgb}{0.5,0.5,0.5}
\definecolor{codepurple}{rgb}{0.58,0,0.82}
\definecolor{mydarkblue}{RGB}{13,39,77}
\definecolor{mypaleblue}{RGB}{0,188,235}
\definecolor{mygreen}{RGB}{116,191,75}
\definecolor{myorange}{RGB}{251,171,24}
\definecolor{darkred}{rgb}{0.5, 0, 0} 
\definecolor{darkgreen}{rgb}{0, 0.5, 0} 
\definecolor{darkblue}{rgb}{0,0,0.5} 
\definecolor{delim}{RGB}{20,105,176}
\definecolor{numb}{RGB}{106, 109, 32}
\definecolor{string}{rgb}{0.64,0.08,0.08}

\lstdefinestyle{mystyle}{
    commentstyle=\color{codegreen},
    keywordstyle=\color{magenta},
    stringstyle=\color{codepurple},
    basicstyle=\ttfamily\footnotesize,
    breakatwhitespace=false,
    breaklines=true,
    captionpos=b,
    frame=single,
    keepspaces=true,
    showspaces=false,
    showstringspaces=false,
    showtabs=false,
    tabsize=2,
    showlines=true
}
\newsavebox{\mylistingbox}

\newcommand{\cmark}{\ding{51}}
\newcommand{\xmark}{\ding{55}}

\newcommand{\sys}{\mbox{Flame}\xspace}

\newcommand{\fig}[1]{Figure~\ref{#1}}
\newcommand{\tref}[1]{Table~\ref{#1}}
\newcommand{\secref}[1]{\S\ref{#1}}
\newcommand{\ra}[1]{\renewcommand{\arraystretch}{#1}}

\newcommand{\mydarkbluedot}{\protect\tikz[baseline=-0.5ex]\protect\draw[black,fill=mydarkblue,radius=3pt] (0,0) circle;}
\newcommand{\mypalebluedot}{\protect\tikz[baseline=-0.5ex]\protect\draw[black,fill=mypaleblue,radius=3pt] (0,0) circle;}
\newcommand{\mygreendot}{\protect\tikz[baseline=-0.5ex]\protect\draw[black,fill=mygreen,radius=3pt] (0,0) circle;}
\newcommand{\myorangedot}{\protect\tikz[baseline=-0.5ex]\protect\draw[black,fill=myorange,radius=3pt] (0,0) circle;}

\newcommand{\hollowcircled}[1]{\tikz[baseline]{%
    \node[anchor=base, draw, circle, inner sep=0, minimum width=0.9em]{#1};}}

\newcommand{\todotemplate}[3]{%
	\mbox{}		
	\marginpar{%
		\colorbox{#2!80!black}{\textcolor{white}{#1}}%
		\vspace*{-20pt}
	}%
	\textcolor{#2}{{#3}}%
}

\newenvironment{tightitemize}%
 {\begin{list}{$\bullet$}{%
 		\setlength{\leftmargin}{10pt}
        \setlength{\itemsep}{0pt}%
        \setlength{\parsep}{0pt}%
        \setlength{\topsep}{0pt}%
        \setlength{\parskip}{0pt}%
        }%
 }%
{\end{list}}
\newcounter{tecounter}
 {\begin{list}{\arabic{tecounter}.}{%
 		\usecounter{tecounter}
 		\setlength{\leftmargin}{10pt}
        \setlength{\itemsep}{0pt}%
        \setlength{\parsep}{0pt}%
        \setlength{\topsep}{0pt}%
        \setlength{\parskip}{0pt}%
        }%
 }%
{\end{list}}%
 {\begin{list}{}{%
 		\setlength{\leftmargin}{10pt}
        \setlength{\itemsep}{0pt}%
        \setlength{\parsep}{0pt}%
        \setlength{\topsep}{0pt}%
        \setlength{\parskip}{0pt}%
        }%
 }%
{\end{list}}%

\iftoggle{editing_mode}{
	\newcommand{\ada}[1]{\todotemplate{ada}{cyan}{#1}}
	\newcommand{\hd}[1]{\todotemplate{hd}{orange}{#1}}
	\newcommand{\mlee}[1]{\todotemplate{mlee}{brown}{#1}}

	\newcommand{\todo}[1]{\textcolor{red}{#1}}
	\newcommand{\fixme}[1]{\textcolor{red}{#1}}
	\newcommand{\fixed}[1]{\textcolor{red}{\st{#1}}}
	\newcommand{\discussion}[1]{\textcolor{orange}{#1}}

	\newcommand{\reviewer}[1]{\textcolor{violet}{#1}}
}{
	\newcommand{\ada}[1]{}
	\newcommand{\hd}[1]{}
	\newcommand{\mlee}[1]{}
	
	\newcommand{\todo}[1]{}
	\newcommand{\fixme}[1]{}
	\newcommand{\fixed}[1]{}
	\newcommand{\discussion}[1]{}
	
	\newcommand{\reviewer}[1]{\textcolor{green}{}} 
}

\lstdefinestyle{yaml}{
     basicstyle=\color{blue}\ttfamily\footnotesize,
     rulecolor=\color{black},
     string=[s]{'}{'},
     stringstyle=\color{blue},
     comment=[l]{:},
     commentstyle=\color{black},
 }

\lstdefinelanguage{json}{
    numberstyle=\small,
    frame=single,
    rulecolor=\color{black},
    showspaces=false,
    showtabs=false,
    breaklines=true,
    postbreak=\raisebox{0ex}[0ex][0ex]{\ensuremath{\color{gray}\hookrightarrow\space}},
    breakatwhitespace=true,
    basicstyle=\ttfamily\small,
    upquote=true,
    morestring=[b]",
    stringstyle=\color{string},
    literate=
     *{0}{{{\color{numb}0}}}{1}
      {1}{{{\color{numb}1}}}{1}
      {2}{{{\color{numb}2}}}{1}
      {3}{{{\color{numb}3}}}{1}
      {4}{{{\color{numb}4}}}{1}
      {5}{{{\color{numb}5}}}{1}
      {6}{{{\color{numb}6}}}{1}
      {7}{{{\color{numb}7}}}{1}
      {8}{{{\color{numb}8}}}{1}
      {9}{{{\color{numb}9}}}{1}
      {\{}{{{\color{delim}{\{}}}}{1}
      {\}}{{{\color{delim}{\}}}}}{1}
      {[}{{{\color{delim}{[}}}}{1}
      {]}{{{\color{delim}{]}}}}{1},
}

\begin{document}

\title[]{\Huge \bf \sys: Simplifying Topology Extension in Federated Learning}

\author{Harshit Daga}
\authornote{Equal contribution}
\affiliation{\institution{Georgia Institute of Technology}}

\author{Jaemin Shin}
\authornote{Work done at Cisco Research}
\affiliation{\institution{KAIST}}

\author{Dhruv Garg}
\affiliation{\institution{Georgia Institute of Technology}}

\author{Ada Gavrilovska}
\affiliation{\institution{Georgia Institute of Technology}}

\author{Myungjin Lee}
\authornotemark[1]
\authornote{Corresponding author}
\affiliation{\institution{Cisco Research}}

\author{Ramana Rao Kompella}
\affiliation{\institution{Cisco Research}}

\renewcommand{\shortauthors}{}

\begin{abstract}
Distributed machine learning  approaches, including a broad class of federated learning (FL)  techniques, present a number of benefits  when deploying machine learning applications over widely distributed infrastructures. The benefits are highly dependent on the details of the underlying machine learning topology, which specifies the functionality executed by the participating nodes, their dependencies and interconnections. Current systems lack the flexibility and extensibility necessary to customize the topology of a machine learning deployment. We present {\bf \sys}, a new system that provides flexibility of the topology configuration of distributed FL applications around the specifics of a particular deployment context, and is easily extensible to support new FL architectures.  \sys achieves this via a new high-level abstraction Topology Abstraction Graphs (TAGs).  TAGs decouple the ML application logic from the underlying deployment details, making it possible to specialize the application deployment with reduced development effort. \sys is released as an open source project, and its flexibility and extensibility support a variety of topologies and mechanisms, and can facilitate the development of new FL methodologies. 
\end{abstract}

\maketitle
\section{Introduction}
\label{s:introduction}

The proliferation of sensors and connected devices such as mobile devices, wearables, and vehicles has resulted in generation of massive amounts of data. In order to quickly and accurately analyze such extraordinarily large and complex data sets to make data-driven decisions, companies have started relying on machine learning techniques. There exist a number of machine learning use cases such as recommendation services~\cite{ml_recommendation_sys_review}; cyber-security breach detection~\cite{ml_cybersecurity_survey}; predictive maintenance and condition monitoring in manufacturing~\cite{ml_manufacturing}; disease identification in healthcare and life sciences~\cite{ml_healthcare_survey} and risk analysis in financial services~\cite{ml_banking_mgt_review}. 

Traditional machine learning approaches require collecting all data together in one place, such as a cloud data center. However, with data sources spread geographically, the network becomes the bottleneck. Additionally, user-privacy laws, such as GDPR~\cite{gdpr}, have resulted in shift towards a federated learning (FL) approach where many clients collectively train a shared model under the orchestration of a central server, also called aggregator.

A classical FL (C-FL) approach adopts a rather simplistic client-server architecture whereby an aggregator (parameter server) builds a global model by combining model updates from training workers (clients). However, not all scenarios fit into this conventional architecture. FL has been a fast-evolving technology and numerous variants~\cite{papaya,fl_at_scale,hierFAVG_topology,joint_hfl,hybrid_fl,vertical_fl} have been proposed. Besides accuracy, these are proposed for different performance objectives such as scalability, convergence, training costs, and so on. The designs are also influenced by factors like operation scales and use cases. Hence, the system architectures are quite different. Some approaches introduce system components like selectors and coordinators as separate runtime entities~\cite{papaya,fl_at_scale}; and others introduce edge aggregators~\cite{hierFAVG_topology,joint_hfl}, enable peer-to-peer collaboration~\cite{p2pfl,canoe}, or take a hybrid approach of combining distributed learning and federated learning~\cite{hybrid_fl}. As a result, one size (i.e., client-server architecture) doesn't fit all.

\begin{figure}[t]
  \centering
  \subfloat[]{\label{fig:topo_dist}\includegraphics[height=0.6in]{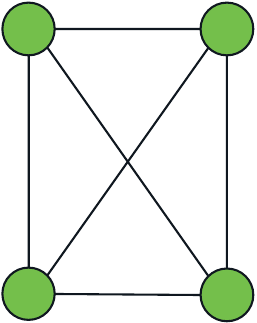}}\hfill
  \subfloat[]{\label{fig:topo_fl}\includegraphics[height=0.6in]{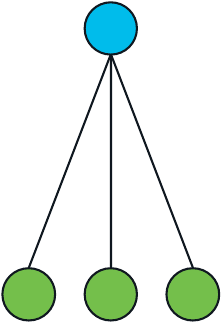}}\hfill
  \subfloat[]{\label{fig:topo_hfl}\includegraphics[height=0.6in]{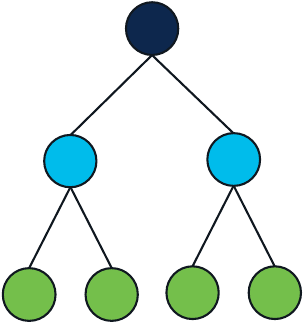}}\hfill
  \subfloat[]{\label{fig:topo_hfl_coord}\includegraphics[height=0.6in]{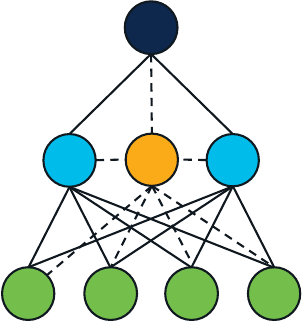}}\hfill
  \subfloat[]{\label{fig:topo_fl_hybrid}\includegraphics[height=0.6in]{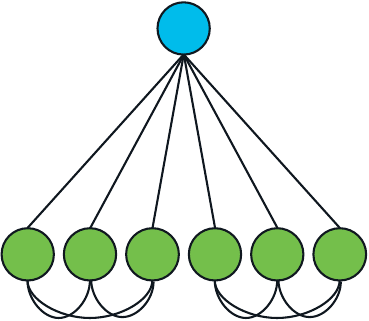}}
  \caption{Topologies that can be used in federated learning: (a) distributed, (b) classical FL, (c) hierarchical, (d) hierarchical with replicas and
    coordinator, and (e) hybrid.  \mygreendot: training node, \mypalebluedot: aggregation node, \mydarkbluedot: global aggregation node, and
    \myorangedot: coordinator. In (d), dotted lines denote connections between coordinator and other components.}
  \label{fig:topo_examples}
\end{figure}

The architecture of the distributed machine learning system defines a specific deployment {\bf topology}. \fig{fig:topo_examples} illustrates several topologies. The topology is defined by the different nodes participating in the machine learning job, the specific functionality they provide (e.g., train, aggregate, select, etc.), their dependencies and the data or control communication channels that interconnect them (e.g., see \fig{fig:topo_hfl_coord}). A topology is further defined by the requirements of its deployment, such as concerning specific communication protocol requirements, or the placement of specific functionality relative to the data it should operate on.

Different topologies introduce benefits for different deployment contexts (in terms of scale or geo-distribution, network connectivity, failures or churn), workload properties (e.g., in terms of update frequency or concept drift), and privacy requirements.
They also present different tradeoffs in terms of training goals such as time-to-convergence or data transfer costs. It is therefore important to enable flexibility when configuring the distributed training topology. This is further amplified by the continued progress in new training architectures which introduce new types of functionality~\cite{oort_selection, fedcs_selection} or cross-participant interactions~\cite{p2pfl}, and opportunities for optimizations that can be realized by updating the topologies of existing systems. 

To make it possible to customize the topology of a machine learning deployment, an FL system needs to be flexible and easily extensible; at the same time, it should be able to decouple the management of learning logic, compute and data.  The flexibility is provided by many existing solutions, but the latter requirements are not fully met. Current frameworks such as FedScale~\cite{fedscale}, Flower~\cite{flower} and PySyft~\cite{pysyft} provide low level APIs which make them flexible. However, they cannot be easily extended to support different deployment scenarios such as hierarchical and hybrid FL, as they lack abstraction suitable for expressing those scenarios.

A recent effort, FedML~\cite{fedml}, offers client-server architecture-based abstraction to improve extensibility. This abstraction provides improved expressiveness compared to other frameworks and enables a few templatized deployments. However, it quickly becomes difficult to support scenarios where FL components don't fit as either client or server. A canonical example is architectures in \cite{papaya,fl_at_scale} where there exist diverse interactions among aggregator, selector and coordinator; classifying them as either client or server gets complicated. Therefore, extending and evolving the deployment scenarios supported by FedML, demands intrusive changes in its codebase, and poses limitations to the flexibility supported by the framework.

In response, we present {\bf \sys}, \textit{a new system that provides flexibility of the topology configuration of distributed FL applications around the specifics of a particular deployment context, and is easily extensible to support new FL architectures and algorithms}. To achieve this, \sys introduces a new abstraction called \textit{Topology Abstraction Graph (TAG)}. This abstraction enables explicit customization of individual components in the system and supports various designs without requiring modifications to the core system components. The higher level {\em TAG} abstraction allows for flexible expression of how these components combine and how they are deployed. \sys also provides interfaces to describe and integrate with different compute infrastructure and dataset providers. This enables support for different resource orchestrators and heterogeneous deployment platforms, as well as to specify and enforce different deployment constraints in terms of data and compute coupling.

The APIs of \sys allow users to express their deployment in a compact TAG representation, provide the machine learning code for the respective roles and select a communication backend. With the APIs, users can extend their use cases easily (\secref{s:extend_topo}). \sys then expands the TAG to map its physical deployment, using information registered with the system about the properties of nodes and datasets. In \secref{s:flexible_backend}, we demonstrate how users can leverage \sys's flexible backend to adapt to their deployment environment or use cases. The abstract representation supported by \sys allows the users to update their topology by merely updating the TAG graph and providing definitions for any new roles or channel protocols, without changes in the core library. Therefore, users can easily switch from one mechanism/topology to another (\secref{subsec:topo_transformation}). In \secref{subsec:sys_comparision}, we compare \sys with other frameworks and showcase that \sys supports a variety of topologies which cannot be easily supported by others. \sys is released as an open source project\footnote{\url{https://github.com/cisco-open/flame}}, and its flexibility and extensibility offer a variety of topologies and mechanisms, and can facilitate the development of new FL methodologies.

\section{Background and Motivation}
\label{section:background_motivation}

\subsection{Federated Learning}
\label{s:fl_intro}

FL~\cite{fedavg} has recently emerged as a compelling machine learning practice for meeting data sovereignty and preserving privacy. There exist many variants of federated learning. Many focus on algorithms~\cite{fedavg, fedprox, fedbuff, qffl, google_adaptive_fl} for improving fairness, accuracy and convergence time. Others propose to select clients intelligently~\cite{oort_selection, fedcs_selection} or to carefully sample a client's dataset~\cite{fedbalancer} for faster convergence. All of these assume the C-FL setting where all clients talk to an aggregation server. In contrast, approaches like hierarchical FL (H-FL)~\cite{hierFAVG_topology, joint_hfl}, hybrid FL~\cite{hybrid_fl} and peer-to-peer FL~\cite{p2pfl} propose different topologies, and thus entail system architectures and communication patterns different from those of the C-FL.

FL can be used in a cross-device manner, where training is done in-situ on devices, and centrally aggregated. Gboard (a virtual keyboard on Android phone) which recommends keywords that user may type on the keyboard~\cite{fl_gboard, fl_gboard2} is a canonical example. It can also be used in a cross-silo manner, where training is performed on distributed infrastructure, near data, but possibly on separate  compute resources where datasets need to be available. Our work considers such cross-silo settings. Its core concept, however, is generic, and thus it can be applied in other FL settings.

\begin{figure}[t]
  \centering
  \subfloat[]{\label{fig:tta_60}\includegraphics[height=1.28in]{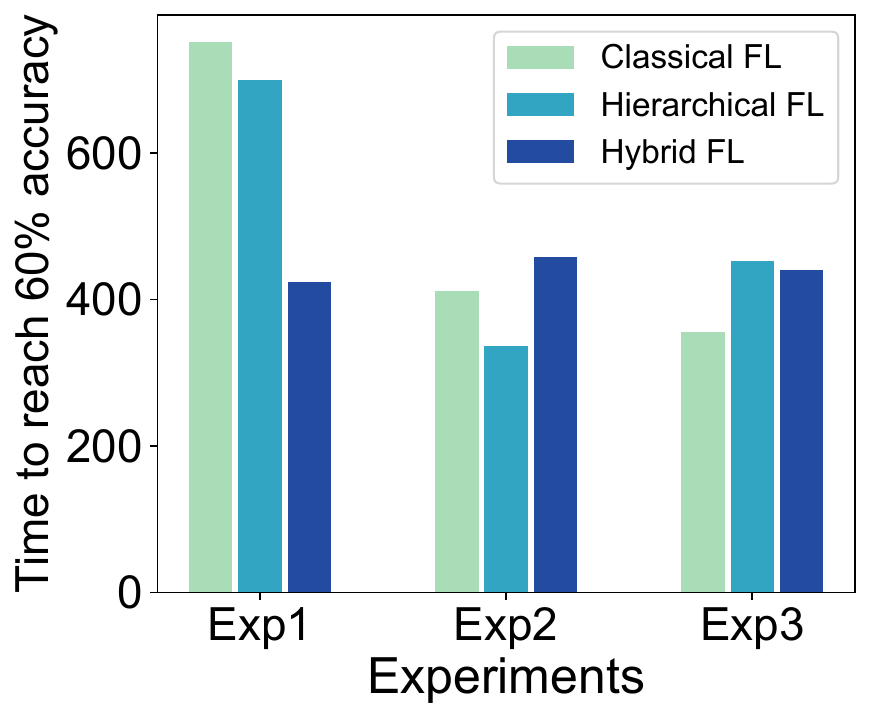}}\hfill
  \subfloat[]{\label{fig:acc_at_1000}\includegraphics[height=1.3in]{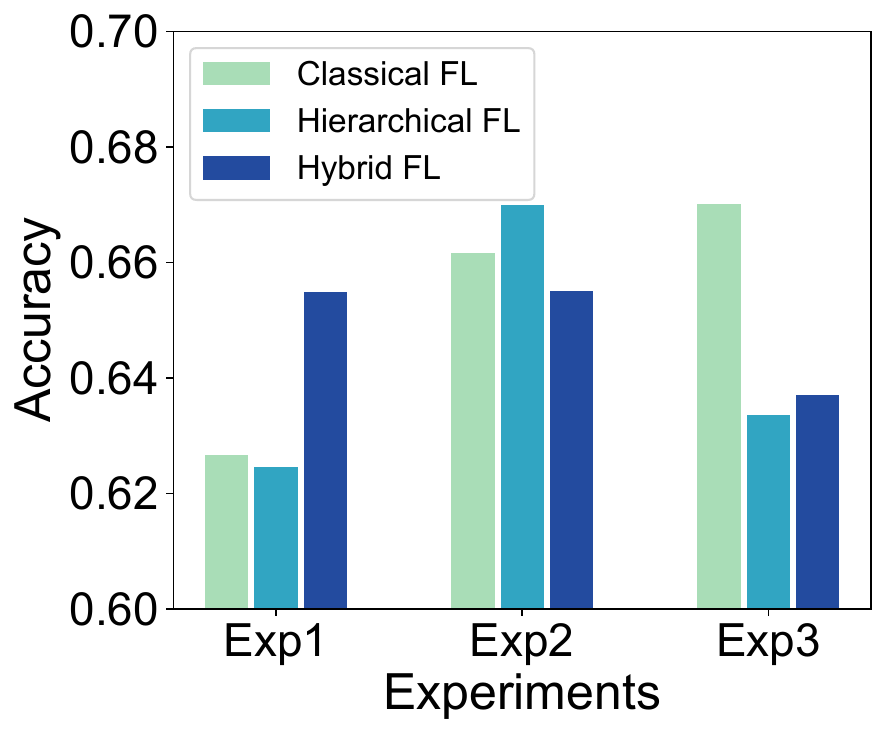}}\hfill
  \caption{Federated learning experiments where each topologies have different time-to-accuracy performance and communication cost: Exp1) straggler node, Exp2) client node failures, and Exp3) aggregator \& leader node failures.}
  \label{fig:topo_motivation_experiments}
\end{figure}

\subsection{Need for Topology Customization}
\label{s:topology}
In FL, the choice of topology can significantly impact the overall efficiency of FL jobs, affecting factors such as time-to-accuracy. Furthermore, the initial conditions assumed for different scenarios may evolve over time, necessitating changes in deployment strategies. These changes can range from simple extensions of a two-level hierarchy to a multi-level hierarchy, or the addition of entirely new components, to modifications in the communication backend.

There are various scenarios that can trigger the need for topology changes in FL. For instance, a one-level C-FL topology may suffice for small-scale experiments. As the experiment scale grows, globally geo-distributed devices favor other topologies, such as hierarchical topologies~\cite{hierFAVG_topology, joint_hfl, demystifying}. In addition, a large-scale hierarchical deployment introduces new components such as selector and coordinator~\cite{fl_at_scale, papaya}, requiring a different topology (e.g., \fig{fig:topo_hfl_coord}).

We demonstrate trade-offs among three topologies: C-FL, H-FL, and hybrid, with respect to accuracy and convergence time using three scenarios with the CIFAR-10 dataset and 50 training nodes. In C-FL, these nodes are connected to a global aggregator (\fig{fig:topo_fl}). For H-FL, the trainers are equally divided into two groups and attached to their corresponding intermediate aggregator (\fig{fig:topo_hfl}). In the hybrid approach, nodes use distributed learning and a leader node in each group shares updates with the global aggregator (\fig{fig:topo_fl_hybrid}).

As shown in the ``Exp1'' of \fig{fig:tta_60}, straggling nodes can cause delays in convergence time where hybrid learning is able to reach $60\%$ accuracy up to 1.77$\times$ faster than C-FL and H-FL. In the event of a training node failure (``Exp2'' from the same figure), H-FL reached $60\%$ accuracy 1.36$\times$ and 1.23$\times$ faster than hybrid and C-FL,  respectively. However, if the intermediate aggregator in H-FL or a group leader in hybrid fails without any failover protection, (shown in ``Exp3'' of \fig{fig:acc_at_1000}), C-FL reaches $67\%$ accuracy whereas H-FL and hybrid topologies achieve only about $63\%$ accuracy. The experiments illustrate that the ``best'' topology depends on factors such as operating conditions or target metrics.

Different topologies can require communication backend changes. For instance, existing frameworks employ different protocols (e.g., MQTT and gRPC) as their communication backend. MQTT can be suitable for C-FL as it simplifies operations (e.g., one firewall rule update and proxy setup) since an MQTT broker only needs to be publicly reachable and other entities such as aggregators and trainers can be hidden from the Internet. On the other hand, protocols like gRPC increase management complexity as firewall and proxy settings may need to be updated for each IP or domain name of the entities (e.g., aggregators). For H-FL and hybrid FL, a single MQTT broker causes communication inefficiency as all the traffic routes through the broker. The capability to configure different communication backends can have a direct impact on the operational overheads. In case of the hybrid topology, device-to-device can use gRPC while device-to-server can leverage MQTT as gRPC does not incur more management overhead for the devices in the same LAN or infrastructure while MQTT still preserves the operational simplicity for connectivity across the Internet.

In summary, when selecting a topology and associated training methods, it is crucial to consider that there is \textit{no single topology that applies to all scenarios}. The C-FL architecture is simple and entails lower management overhead but lacks scalability. Hierarchical topologies offer scalability but introduce the risk of failure in intermediate aggregators, which can result in the loss of local updates from multiple training nodes, impacting accuracy and time-to-accuracy. Similarly, hybrid topologies, despite their communication efficiency, carry the risk of failure due to device failures affecting per-group model aggregation when using communication collectives like ring-all-reduce. It is essential to carefully consider the desired goals and trade-offs when customizing the FL topology. To support such flexibility and extensibility there is a need for systems that provide modular support for different components and communication backends.

\begin{figure}[t]
  \centering
  \includegraphics[scale=0.45]{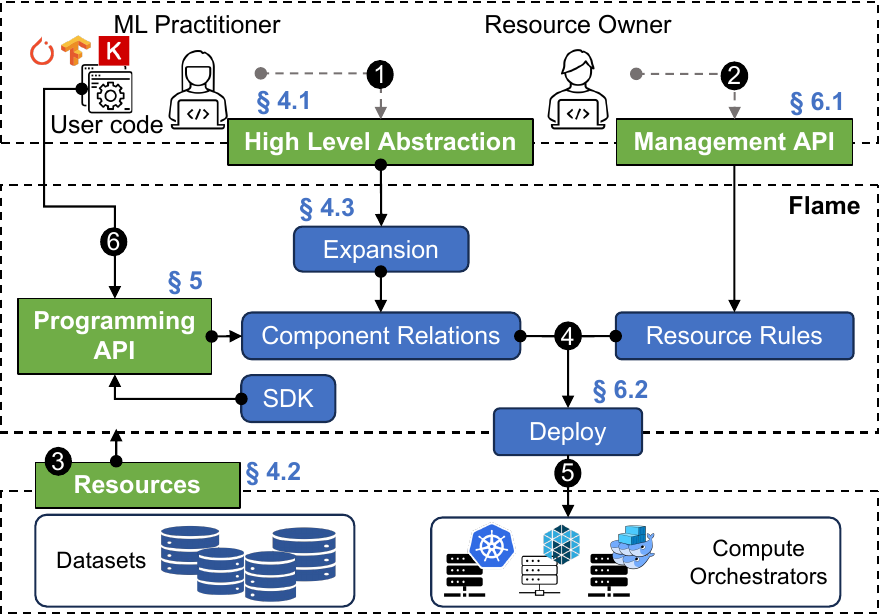}
  \caption{Conceptual overview of the \sys system.}
  \label{fig:sys_overview}
\end{figure}

\subsection{Current Ecosystem}
\label{s:solutions}

To create a machine learning system, frameworks such as PyTorch and TensorFlow, provide APIs and low level support for a wide range of models, with focus on speed and optimization. However, creating and deploying such models in geo-distributed settings requires support from resource orchestrators, model registry to store model snapshots and integration with monitoring tools. Alternatively, platforms such as Amazon Sage Maker, Azure ML~\cite{aws_fl}, Vertex AI, offer means for deploying ML applications in the cloud, with a goal to streamline the operation of ML jobs. However, there is no emphasis on support for programmatically extending and adapting the details of the topology of the deployed jobs.

FedML~\cite{fedml} is a recent framework that aims to support easy development of FL-based ML models. Its API and client-server based abstraction allow development of new scenarios, thereby assisting few requirements discussed in this paper. However, the client-server abstraction is not enough to make FedML easily extensible. In FedML, a node is either client or server. Consider the topology shown in \fig{fig:topo_hfl} designed for hierarchical FL. While training node and global aggregation node fit well with the client-server abstraction, intermediate aggregators don't, because they act as both client as well as server depending on which component they interact with. In order to handle this dilemma, FedML introduces a concept called \textit{rank} and, based on the rank's value, implements different behaviors in its client codebase. While this enables support for H-FL, it is a stop gap. In hierarchical FL with a separate coordinator (\fig{fig:topo_hfl_coord}), the rank value can't help because it is unclear which value to assign to rank. 

Moreover, while topologies for classical FL (\fig{fig:topo_fl}) and hybrid FL (\fig{fig:topo_fl_hybrid}) look similar, behaviors of training and aggregation nodes are dissimilar even though trainers in both topologies are classified as client. Hence, classifying a role as client or server is too coarse to support emerging and diverse FL scenarios. This limitation can  require intrusive source code changes in the core codebase. Other frameworks such as FedScale~\cite{fedscale}, Flower~\cite{flower} and PySyft~\cite{pysyft} follow the same client-server architecture or two-tier topology, and share the same shortcomings of FedML in terms of extensibility.

\section{\sys Overview}
\label{section:overview}

\noindent{\bf Goal.} The primary objective of \sys is to offer a fine-grained abstraction that eases the composition and extension of machine learning topologies. The system aims to provide modular building blocks that facilitate swift integration and experimentation with new components, new deployment strategies and topologies. \sys also provides precise control over the communication backend to reduce management overheads of FL jobs under different topologies.

\noindent{\bf Need for New System.}
The discussion in the previous section points to two key challenges and limitations associated with existing approaches. We summarize them as follows.

\textit{C1:} There is a lack of higher-level, modularized representation of a machine learning application, that explicitly describes all of its components and their dependencies, and could therefore make it possible to flexibly adapt and specialize their behavior and deployment details. 

\textit{C2:} In a geo-distributed environment, deploying FL jobs
requires coordinating across distinct data and infrastructure providers, possibly owned by different entities. To provide deployment flexibility, there is a need to decouple the infrastructure dependency from the machine learning tasks, yet to provide sufficient meta information that will support mapping and configuration decisions. 

\noindent{\bf Overview.} The system concept of \sys, as illustrated in \autoref{fig:sys_overview}, revolves around three key components: a high-level abstract description of machine learning job, resource descriptions, and APIs (programming and management) for extending learning techniques and managing jobs. First, a complex machine learning job is described by the developer in the form of \textit{Topology Abstraction Graph} (TAG) (\secref{subsection:tag} and \hollowcircled{1}). TAG employs a graph-based representation that maps the expanded physical topology to a condensed logical structure, in a manner that captures the functional characteristics of each components and their deployment constraints. Each node in the TAG represents a {\em role} abstraction, and roles are interconnected using {\em channels}. Using a graph-based representation, over a client-server architecture, provides  expressiveness and extensibility benefits: (i) expressing the responsibility of any component as independent roles; (ii) breaking the components into individual roles provides us with fine-grained abstraction for communication backend between the roles which can be easily switched without impacting the machine learning logic and; (iii) all topologies can be expressed as a graph and TAG can abstract a physical deployment. 

\sys is designed to operate on diverse distributed resources (infrastructures and datasets) provided by different entities. These get integrated with \sys through the services provided by the {\em management plane} (\secref{s:components} and \hollowcircled{2}). During registration, the resources provide metadata attributes that describe their properties and access constraints. This  provides application developers with the flexibility to choose resources based on their specific requirements (\secref{s:infra_mgt} and \hollowcircled{3}).

The abstract representation of the TAG is expanded to identify the various components and their relationships (\secref{s:expansion} and \hollowcircled{4}). Following the expansion of the TAG into a tangible physical instantiation, \sys leverages the resource metadata and the user-defined infrastructure constraints to facilitate specific deployment decisions (\secref{s:workflow} and \hollowcircled{5}). Finally, \sys offers support for diverse topologies and provides APIs through a {\em programming model} to enhance extensibility (\secref{s:prog_model} and \hollowcircled{6}). These APIs empower developers to develop novel learning techniques or modify existing ones to support different learning techniques associated with distinct roles. 

\section{Design}
\label{s:design}

\begin{figure}[t]
  \centering
  \subfloat[]{\label{fig:tag}\includegraphics[height=0.5in]{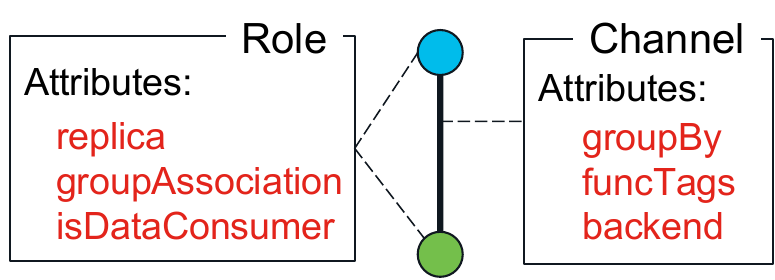}}\hfill
  \subfloat[]{\label{fig:topo_tag_dist}\includegraphics[height=0.5in]{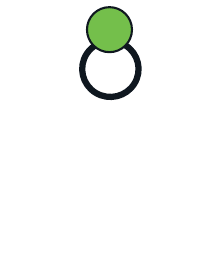}}\hfill
  \subfloat[]{\label{fig:topo_tag_fl}\includegraphics[height=0.5in]{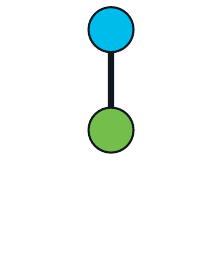}}\hfill
  \subfloat[]{\label{fig:topo_tag_hfl}\includegraphics[height=0.5in]{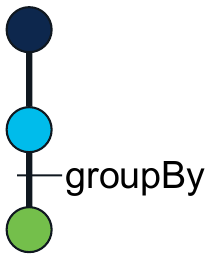}}\hfill
  \subfloat[]{\label{fig:topo_tag_fl_hybrid}\includegraphics[height=0.5in]{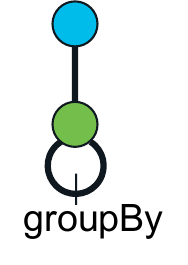}}
  \caption{(a) Building blocks of TAG. TAG representation of topologies: (b) distributed, (c) classical FL, (d) hierarchical FL, (e) hybrid. \mygreendot {} is a trainer node with isDataConsumer set. The groupBy attribute is used to create groups of the same role.
  }
  \label{fig:topo_tag_examples}
\end{figure}
\begin{figure*}[t]
  \centering
  \subfloat[]{\label{fig:tag_expand1}\includegraphics[height=1.53in]{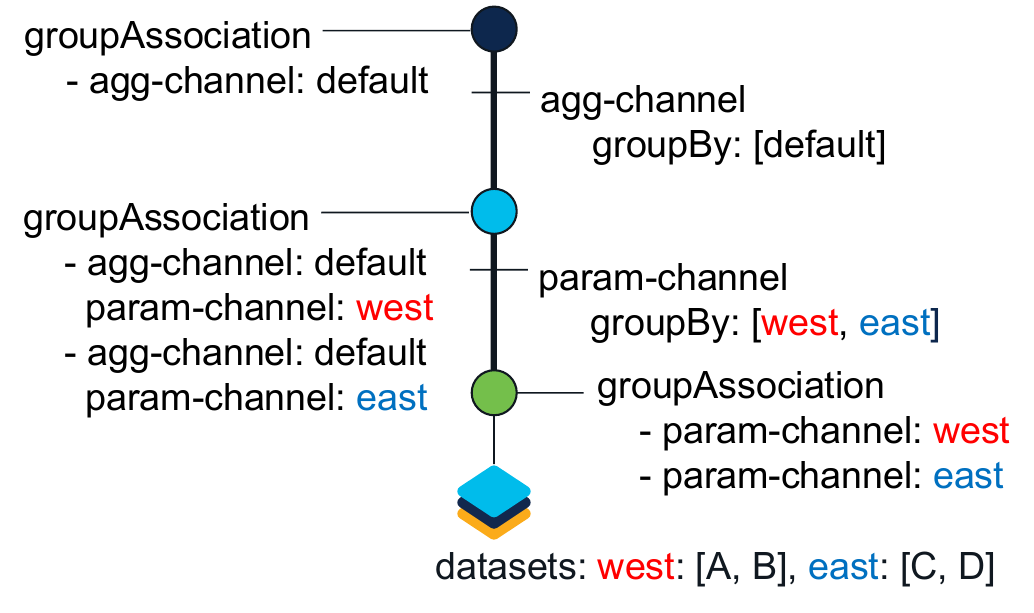}}
  \subfloat[]{\label{fig:tag_expand2}\includegraphics[height=1.53in]{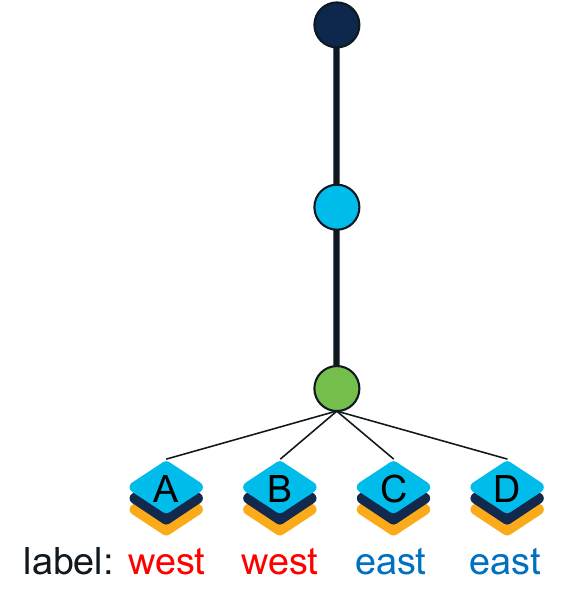}}
  \subfloat[]{\label{fig:tag_expand3}\includegraphics[height=1.53in]{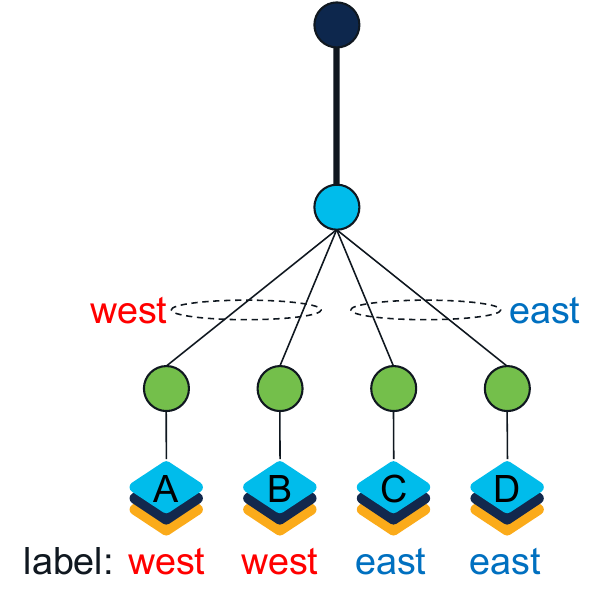}}
  \subfloat[]{\label{fig:tag_expand4}\includegraphics[height=1.53in]{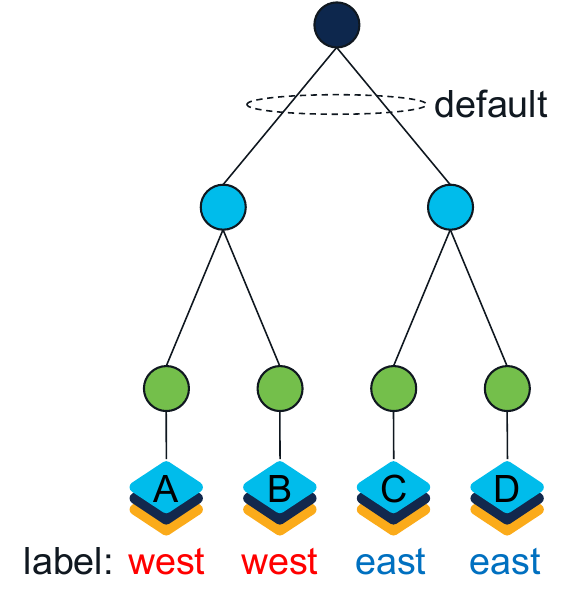}}
  \caption{Expansion of a TAG into a physical topology. (a) TAG representation of hierarchical FL (H-FL); each dataset belongs to a group; (b) dataset is expanded; (c)
  one training worker is allocated per dataset and the worker's group is determined by the dataset's group; and (d) each entry in \texttt{groupAssociation}
  corresponds to one aggregation worker; hence, two aggregators are created; each belongs to a different group for param-channel while both have
  one ``default'' group for agg-channel. Since there is one entry for \texttt{groupAssociation} in the global aggregation role, expansion finishes by
  creating one global aggregation worker. Note that the expansion of roles can be done in an arbitrary order since \texttt{groupAssociation} has all
  the necessary information for expansion. A channel's attributes (e.g., \textit{groupBy}) are used to validate the expansion.}
  \label{fig:tag_expand}
\end{figure*}

\subsection{Topology Abstraction Graph}
\label{subsection:tag}

A central abstraction in \sys is \textit{Topology Abstraction Graph} (\textbf{TAG}). It represents a simple logical graph, which allows users to express a training workload declaratively for any ML job. It comprises two basic building blocks: \textit{role} and \textit{channel}. A role is a vertex and serves as an abstraction for worker while a channel is an undirected edge between a pair of roles and acts as an abstraction for communication backend. TAG's schema is visually represented as illustrated in \fig{fig:tag} and different learning topologies can be represented using a TAG as shown in Figures~\ref{fig:topo_tag_dist}-\ref{fig:topo_tag_fl_hybrid}. Below we discuss role and channel in detail. To aid understanding of the discussion, a concrete example of a TAG is presented in \fig{fig:tag_expand1}.

\noindent\textbf{Role.} An executable worker unit carrying out a specific task in an ML job is defined as a role. Depending on the topology, the task and behavior of a role can vary. For instance, a training worker in FL uses data to build a local model and sends model updates to an aggregation worker, which combines them to create a global model. In hierarchical FL, the aggregator may forward the aggregated model to a global aggregation worker. By exploiting the  uniqueness of the tasks associated with these workers, \sys is able to abstract their behavior in the learning process and assigns them roles. This forms the foundation of \sys's flexible and extensible system. These roles are associated with programs that are defined at the job composition stage. The programs are made up of a set of functions such as train, evaluate, load data, and get/distribute model updates, based on the role's responsibility. The program also contains information about the functions execution, as described in Section~\ref{s:prog_model}.

The flexible binding between role and program allows \sys to be extensible and support different mechanisms under different topologies. In order to express the specific requirements associated with the deployment of a worker, each role has three attributes: \textit{isDataConsumer}, \textit{groupAssociation} and \textit{replica}.  A TAG is expanded into its physical topology by leveraging these attributes, as discussed in \secref{s:expansion}. In order to distinguish training nodes from other components, the boolean attribute \textit{isDataConsumer}, is used to indicate whether a particular role consumes data or not.

In case of topologies like H-FL and hybrid, which involve grouping different nodes to form a cluster, the association of a worker instance (of a role) with channels and their respective groups is determined by the \textit{groupAssociation} attribute (more on channel below). Workers from different roles are connected through channel and the \textit{groupAssociation} governs the connectivity between workers of the same role and those belonging to other roles. This attribute contains a list of the following set: \{$c_1$:$g_1$, $...$, $c_i$:$g_i$\} where $c_i$ is the name of channel $i$ and $g_i$ is a group in the channel; an example of this attribute is shown in \fig{fig:tag_expand1}. The size of the list corresponds to the number of workers for the given role.

Finally, to ensure failure protection or load-balancing, the \textit{replica} attribute is provided to determine the number of workers (i.e., the number of instances) assigned to a particular role. These replicated workers possess identical properties and configurations, and conduct the same type of work. The way they are coordinated however determines the amount of work in each instance. This attribute is particularly valuable for distributing aggregation work among aggregators.

\noindent\textbf{Channel.} 
It is an abstraction that links a pair of roles in the TAG and facilitates the exchange of data between them through a communication channel. This design choice enables \sys to offer precise control over the communication backend used for each channel, by specifying the desired communication protocol or messaging service. This is in contrast to other systems which allow for configuration of a single backend across the entire ML job, and facilitates the design of efficient ML jobs tailored to the user's specific requirements and resource availability. 

Channel has three key attributes: \textit{groupBy}, \textit{funcTags} and \textit{backend}. The \textit{groupBy} attribute is responsible for grouping roles that are connected through the channel. Currently, \sys utilizes a label-based approach, but the attribute can be easily extended to support customized grouping algorithms.

The \textit{funcTags} attribute maps the end-points of the channel to the functions within the connected roles. The attribute maintains key-value pairs where key is a role and value is a list of functions exposed by the application code associated with the role. For instance, in the channel between trainer and aggregator, the \textit{funcTags} has ``fetch'' and ``upload'' functions for the training role. Similarly, the attribute has ``distribute'' and ``aggregate'' functions for the aggregation role. This attribute helps avoid any ambiguity in identifying which functions to execute on a specific channel when a role is connected to multiple channels.

The \textit{backend} attribute is used to determine the communication protocol for a channel. Users may choose to store their datasets in the cloud of one or multiple providers. Or, they may choose to keep their datasets across different regions of the same provider's cloud. In such cases, co-location of datasets naturally leads to co-location of trainers; and using one type of backend may result in inefficiency (e.g., MQTT traffic over WAN via a broker), increased complexity (e.g., multi-broker setting for MQTT or complex configuration updates for firewall, ACL and reverse proxy in case of non-broker communication protocol such as gRPC) or both. By allowing per-channel backend, these limitations can be mitigated. We show the utility of this attribute in \secref{s:flexible_backend}.

\subsection{Resource Annotation for Deployment}
\label{s:infra_mgt}
In an FL job, resources such as compute and dataset are required but may not be owned by the user deploying their learning tasks. To enable deployment of FL jobs under such situations, \sys allows 
resource owners to independently register and annotate their resources, which can then be used as needed by the learning job. This approach effectively decouples the infrastructure dependency from any learning tasks and provides greater flexibility in resource usage.

\noindent\textbf{Compute Access.}
Most of the current systems assume that compute nodes are managed through a single provider or utilize a single cluster management tool. However, \sys distinguishes itself by providing an integration service that supports various resource orchestration managers and allows users to register their own cluster (\secref{s:components}). During the compute registration phase \sys receives detailed information about computing clusters, including geographical boundaries, resource capacity, and resource types. This information can then be utilized by users when submitting ML jobs or by dataset owners to establish accessibility boundaries for their workers or datasets respectively. 

\noindent\textbf{Metadata for dataset.}
For ML training, datasets need to be associated with a configuration for training workers to consume. \sys requires data owners to independently register metadata information with the system, which includes the \textit{realm} and \textit{url} of the dataset. The \textit{realm} attribute 
plays a crucial role in defining access restrictions for the dataset. For instance, in order to comply with data privacy regulations like GDPR~\cite{gdpr}, the dataset owner can utilize information on cluster geographical boundaries to restrict accessibility to specific regions. This design ensures compliance with security and privacy policies associated with the dataset, empowering data owners to maintain control over its accessibility and deployment while allowing others to utilize it. It is important to note that \sys only stores metadata information, not the raw dataset whose location is pointed by the \textit{url}. Upon registering a dataset, \sys assigns a unique \textit{metadataID} to it, which is used by the users when submitting ML jobs (\secref{s:expansion}). When users provide a job specification, they can use a \textit{datasetGroups} attribute to combine different datasets into separate groups. This is then used by \sys to map and incorporate the dataset information prior to TAG expansion. To illustrate this, consider the example in \fig{fig:tag_expand1}, where a user has formed two groups "west" and "east", each consisting of two distinct datasets: (A, B) and (C, D), respectively.

\begin{algorithm}[t]
  \caption{TAG expansion}
  \label{alg:tag_expand}
  \small
  \DontPrintSemicolon
  \SetKwInOut{Input}{Input}\SetKwInOut{Output}{Output}
  \SetKwProg{Fn}{Function}{:}{}
  \SetKwFunction{FExp}{Expand}
  \SetKwFunction{FBuild}{BuildWorkers}
  \SetKw{Continue}{continue}

  \Fn{\FExp{$J$}}{
    \tcp{$J$: job specification}
    $W \gets \phi$ \tcp{$W$: a total list of workers}
    
    \If{\texttt{PreCheck($J$)} is false}{
      \KwRet $\phi$
    }
    
    $R \gets$ \texttt{GetRoles($J$)} \tcp{$R$: roles}
    \For{$r \in R$}{\label{alg:exp_start}
      $X \gets$ \texttt{BuildWorkers($r$, $J$)}\; 
      $W \gets W \cup X$\;
    }

    \If{\texttt{PostCheck($W$, $J$)} is false}{
      \KwRet $\phi$
    }
    \KwRet $W$
  }
  \Fn{\FBuild{$r$, $J$}}{
    $W \gets \phi$\;
    \eIf{$r$ is a data consumer}{
      $G \gets$ \texttt{GetGroupsOfDataSets($r$, $J$)}\; \label{alg:exp_data_config_start}
      \For{$g \in G$}{
        $D \gets$ \texttt{GetDataSets($g$)}\;
        \For{$d \in D$}{
          $m \gets$ \texttt{GetComputeId($d$)}\;
          $a \gets$ \texttt{GetGroupAssocByGroupName($r$, $g$)}\;
          $w \gets$ \texttt{CreateWorkerConfig($r$, $m$, $a$)}\;
          $W \gets W \cup \{w\}$\; \label{alg:exp_data_config_end}
        }
      }
    }{
      $GA \gets$ \texttt{GetGroupAssociations($r$)}\; \label{alg:exp_groupby_config_start}
      \For{$a \in GA$}{
        $c \gets$ \texttt{GetReplicaNum($r$)}\;
        \For{$i = 0; i < c; i++$}{ \label{alg:replica}
          $m \gets$ \texttt{DecideComputeId($a$)}\;
          $w \gets$ \texttt{CreateWorkerConfig($r$, $m$, $a$)}\;
          $W \gets W \cup \{w\}$\; \label{alg:exp_groupby_config_end}
        }
      }
    }
    \KwRet $W$
  }
\end{algorithm}

\vspace{0.3em}
\noindent As part of our design, we deliberately separated the infrastructure from the programming logic to facilitate a more organized approach to managing an FL job. By doing so, users can focus solely on composing their ML job, without worrying about the coupling between compute node and dataset. Without this design choice, developers would be unable to complete the composition of an FL job until a data owner makes a dataset available on a compute node. This "human-in-the-loop" model would significantly slow down the composition and deployment of an FL job. \sys, on the other hand, allows for the automatic acquisition of a compute node and access to a dataset, streamlining the process.

\subsection{TAG Expansion}
\label{s:expansion}
\noindent  Algorithm~\ref{alg:tag_expand} shows the TAG expansion pseudocode. The algorithm expands the abstract representation into a physical deployment topology by creating workers based on the specifications in roles and channels. 

The top-level function, \texttt{Expand} walks through roles (line~\ref{alg:exp_start}) and calls \texttt{BuildWorkers} for each role. Then, \texttt{BuildWorkers} creates the worker configuration. The specification for each role is self-contained. Thus, there is no particular order to iterate roles. If a role is a data consumer, the function iterates on datasets for the role, creates one worker configuration per dataset (lines~\ref{alg:exp_data_config_start}-\ref{alg:exp_data_config_end}) and uses the \textit{datasetGroups} to determine the group. Otherwise, the function takes \textit{groupAssociation} values of role $r$ and creates the corresponding worker (lines~\ref{alg:exp_groupby_config_start}-\ref{alg:exp_groupby_config_end}). During the expansion, if \textit{replica} is set for a role (not a data consumer), the algorithm creates copies of the role (line~\ref{alg:replica}). Those copies share the same properties (e.g., channel's group). For instance, the topology shown in \fig{fig:topo_hfl_coord} is implemented by using \textit{replica}. The channel information is used in pre and post checks to validate the correctness of the TAG and expanded physical deployment.

\noindent\textbf{Example.} \fig{fig:tag_expand} demonstrates the application of Algorithm~\ref{alg:tag_expand} to expand the high level TAG for H-FL (\fig{fig:tag_expand1}) to a physical deployment topology. To begin, we associate all datasets in the job specification with the trainer (data consumer) role, resulting in one worker per dataset, as shown in \fig{fig:tag_expand2}. Then we compare the values of \textit{datasetGroups} and \textit{groupAssociation} to group the training workers into ``west'' and ``east'' groups. The next step is to use \textit{replica} and \textit{groupAssociation} associated with the ``param-channel'' to determine the number of workers required for the aggregator role. By default, \textit{replica} is set to one, unless explicitly stated. In the example, two aggregation workers are created based on the \textit{groupAssociation} values. The same process is applied for the top-level role (global aggregator). Since there is only one value (i.e., \textit{default}) in the \textit{groupAssociation} attribute, a single worker instance is created (\fig{fig:tag_expand4}). 

\begin{figure}[t!]
  \centering
  \includegraphics[width=0.95\columnwidth]{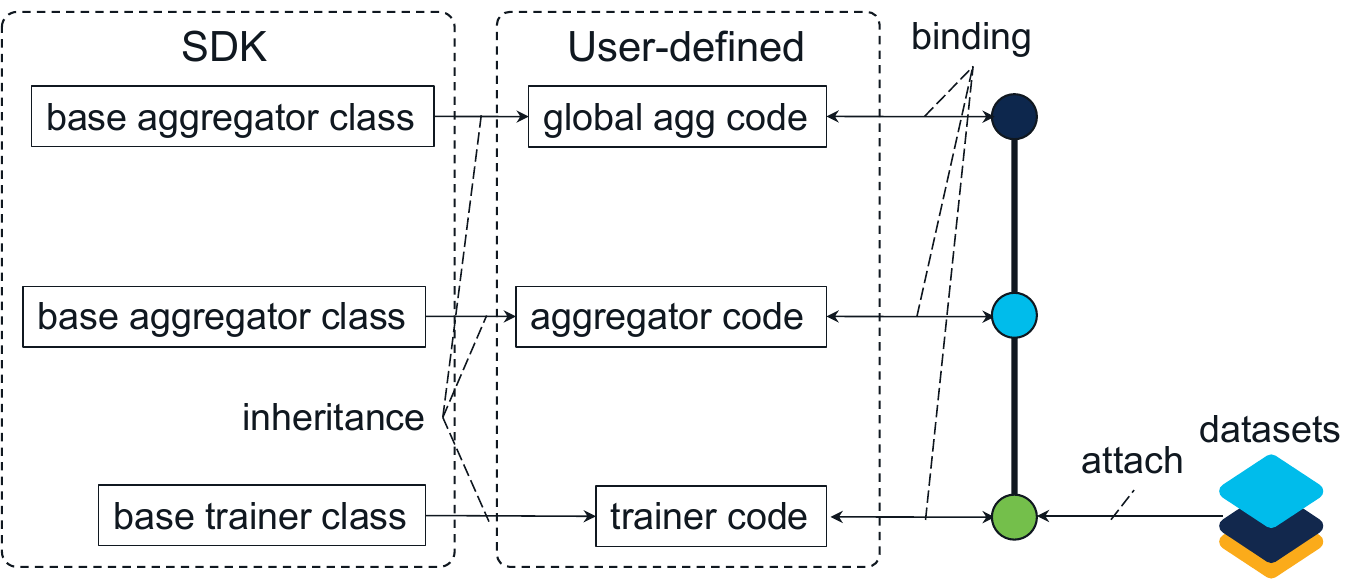}
  \caption{Workload composition for hierarchical topology.}
  \label{fig:tag_hier_ex2}
\end{figure}

\section{Programming Model}
\label{s:prog_model}
\sys provides two programming models: (1) \textit{user} and (2) \textit{developer}. The \textit{user} programming model is for end users who wish to use the \textit{out-of-the-box} functionalities of \sys to deploy a distributed ML job. The \textit{developer} programming model is intended for developers who want to extend the capabilities of \sys by allowing for different topologies, roles, and training methodologies. The former is useful for those whose needs can be met by the built-in functionalities of \sys while the latter is essential when the built-in features are not sufficient to fulfill the user's needs. Therefore, we refer to users or participants as those who mostly rely on the user programming model whereas we denote developers as those who need more than the built-in features.

\begin{figure}[t]
\begin{lstlisting}[frame=tb,language=Python]
from flame.mode.horizontal.trainer import Trainer
class MNistTrainer(Trainer):
  def initialize(self) -> None: 
    # Initialize the model 
  def load_data(self) -> None:
    # Describe operation to handle data
  def train(self) -> None:
    # Training code
  def evaluate(self) -> None:
    # Testing code
t = MNistTrainer(config)
t.compose()
t.run()
\end{lstlisting}
\caption{Code snippet of user-defined MNistTrainer role to illustrate user programming model. After inheriting a base class (\texttt{Trainer}), user only implements four basic functions: \texttt{initialize()}, \texttt{load\_data()}, \texttt{train()}, and \texttt{evaluate()}.}
\label{fig:user_prog_model}
\end{figure}

\noindent\textbf{User Programming Model.} The \sys SDK provides a set of base programs (as Python classes). A user builds a job-specific program by implementing a few core functions (e.g., \texttt{initialize}, \texttt{train}, \texttt{evaluate}, etc). The example shown in \fig{fig:tag_hier_ex2} illustrates the relationship between programs in the \sys SDK and user-defined ones. A user can build the logic for a given role for standard training methodology by inheriting the pre-defined base classes. The base class provides a basic workflow for a certain role (such as trainer, intermediate and global aggregator), allowing the user to focus on implementing relevant core functions for their learning task. For instance, for a hierarchical FL (H-FL) topology user can define their custom MNistTrainer as illustrated in \fig{fig:user_prog_model}, by inheriting the \textit{out-of-the-box} base class provided the SDK. Implementing the aggregator role is simpler than the trainer role as \sys's core library includes essential functions like \texttt{distribute} and \texttt{aggregate}. If aggregator roles need to perform validation tests, users can additionally implement the \texttt{load\_data} and \texttt{evaluate} functions.

\begin{figure}[t]
  \begin{lstlisting}[language=Python,frame=tb]
class Trainer(Role, metaclass=ABCMeta):
  def compose(self) -> None:
    with Composer() as composer:
      self.composer = composer
      tl_load = Tasklet("load", self.load_data)
      tl_init = Tasklet("init", self.initialize)
      tl_train = Tasklet("train", self.train)
      ... ...
      tl_copy = Tasklet("snapshot", self.snapshot)
      loop = Loop(loop_check_fn=lambda: self._work_done)
      tl_load >> tl_init >> loop(tl_get >> tl_train >> ... >> tl_copy >> ...)
\end{lstlisting}
\caption{Code snippet that illustrates a composer mechanism via developer programing model. \texttt{Tasklet} accepts \textit{alias} as the first argument to ease the modification of a tasklet chain.}
\label{fig:dev_prog_model}
\end{figure}

\noindent\textbf{Developer Programming Model.}
\sys is designed to provide extensibility to support different FL topologies. To achieve this, \sys allows developers to extend or create different roles and accommodate other state-of-the-art learning approaches. Internally, each worker executes the functions in the program associated with its role. In \sys, those functions are specified as an execution unit called \texttt{tasklet\footnote{It is to imply that the execution unit is small; it's not one in Linux kernel.}}. Tasklets are combined together to finish a worker's task. Inspired by workflow management solutions~\cite{airflow}, \sys offers functionality to structure a worker's task as a collection of tasklets and present it as a workflow. Since an ML job typically consists of repeating tasklets, the workflow-like approach helps to formalize the development process of any ML mechanisms, thereby facilitating fast development. To create a workflow, \sys overrides the right shift ($\gg$) operator and provides a composer so that various tasklets can be chained together. An additional \texttt{Loop} primitive, allows repeated execution of chained tasklets until an exit condition is met. This methodology provides easy extensibility for a developer to create standalone tasklets such as taking a snapshot of the model, as shown in \fig{fig:dev_prog_model}, or to record various metrics after each step and link them in the workflow.

In addition to its core features, \sys offers a convenient set of API functions through the \texttt{composer} and \texttt{tasklet} modules, which are detailed in \tref{tbl:prog_api}. These APIs enable developers to make surgical modifications to the tasklet chain and to quickly develop new functionalities. With class inheritance, the need to re-chain all the tasklets in the child class is avoided, and only a new tasklet is required for the new functionality. This approach reduces redundant lines of code, avoids core library changes, and reduces the risk of introducing bugs, as shown in \secref{s:extend_topo}.

Finally, in order to ensure compatibility with various communication backends, such as MPI, MQTT, Kafka, and gRPC, \sys's SDK separates the ML logic from the communication layer. This is achieved by providing a \textit{channel manager} interface with a standardized set of APIs, as exemplified in \tref{tbl:ch_api}, which can be utilized by any two connected roles. This abstracted interface for the communication backend allows roles to send and receive messages uniformly regardless of the underlying protocol. Consequently, this not only enhances the flexibility of \sys, but also provides developers with a consistent means of interacting with the system, irrespective of the chosen communication backend.

\begin{table}[t!]
  \small
  \centering
  \ra{1.2}
  \begin{tabular}{@{} l @{} l @{ } >{\arraybackslash}p{1.5in} @{}}
    \toprule
    \textbf{Function} & \textbf{Module} & {\centering\textbf{Note}} \\
    \midrule
    \texttt{get\_tasklet($alisas$)} & composer & Return a tasklet of $alias$ \\
    \texttt{insert\_before($tasklet$)} & tasklet & Insert $tasklet$ before a tasklet \\
    \texttt{insert\_after($tasklet$)} & tasklet & Insert $tasklet$ after a tasklet \\
    \texttt{replace\_with($tasklet$)} & tasklet & Replace $tasklet$ with a tasklet \\
    \texttt{remove()} & tasklet & Remove itself from a chain \\
    \bottomrule
  \end{tabular}
  \caption{Composer and Tasklet API.}
  \label{tbl:prog_api}
 \end{table}

\begin{table}[t]
  \small
  \centering
  \ra{1.2}
  \begin{tabular}{@{} l @{ } >{\arraybackslash}p{2.3in} @{}}
    \toprule
    \textbf{Function} & {\centering\textbf{Note}} \\
    \midrule
    \texttt{join()} & Join channel and allocate resources for the channel \\
    \texttt{leave()} & Leave channel and deallocate its resources \\
    \texttt{send($end$, $msg$)} & Send $msg$ to $end$ \\
    \texttt{recv($end$)} & Receive a message from $end$ \\
    \texttt{recv\_fifo($ends$, $k$)} & Receive a message from the first $k$ ends in a list of $ends$ in a FIFO manner \\
    \texttt{peek($end$)} & Peek a message from $end$ \\
    \texttt{broadcast($msg$)} & Broadcast $msg$ to all the peers at the other end of channel \\
    \texttt{ends()} & Return a list of peers at the other end of channel filtered by a chosen peer selection logic \\
    \texttt{empty()} & Check if peers exist at the other end of channel \\
    \bottomrule
  \end{tabular}
  \caption{Channel API.}
  \label{tbl:ch_api}
 \end{table}


\section{Management Plane} 
\label{s:management}
The management plane of \sys is responsible for managing the lifecycle of FL jobs. Users can create, update, submit and monitor their jobs via a REST API in the management plane. The complete specification can be found in our repository\footnote{\url{https://github.com/cisco-open/flame/tree/main/api}}.

\subsection{System Components}
\label{s:components}
The management plane consists of the following components: APIserver, controller, notifier, deployer, and agent.

\noindent{\bf APIserver.} The APIserver is a front end that exposes the REST API. A CLI tool uses the REST API and allows users to interact with the management plane.

\noindent{\bf Controller.} The controller is the core unit in the management plane. It has three primary responsibilities. (i) It processes requests from users and other system components (e.g., agent and deployer) and manages the state via a database. (ii) It performs TAG expansion into a real topology, and interacts with compute cluster managers, such as Kubernetes, for worker deployment and compute resource provisioning/decommissioning. (iii) Finally, it monitors the job's progress and events such as worker failure.

\noindent{\bf Deployer.} The deployer is an integration interface service, which provides abstraction to integrate different compute orchestration solutions such as Kubernetes, Docker Swarm, Apache Mesos, etc. By implementing the APIs defined in the deployer's interface, \sys can interact with any compute orchestrator. In each compute cluster, the deployer can generate requests for resource allocation and instance (typically in the form of a container) creation, based on instructions received from the controller.

\noindent{\bf Agent.} A learning job in \sys consists of multiple roles executing tasks. Each instance of a role in a cluster includes a thin client called agent. The agent is responsible for managing the lifecycle of a task in a given job via the APIserver. The agent fetches the ML code associated with the role, the channel configuration and meta information on the dataset, all of which are needed by a worker to carry out a task. Once obtained, the agent starts the training by executing a worker role as a child process. It monitors the worker's status and regularly informs the controller. The agent also fulfills the controller's commands such as task termination.

\noindent{\bf Notifier.} 
It is a service that allows the controller to push events 
to agents and deployers. The notifier enables event-driven management of FL jobs because the agents and deployers maintain an active connection with the notifier. 

\begin{figure}[t]
    \centering
    \includegraphics[width=0.88\columnwidth]{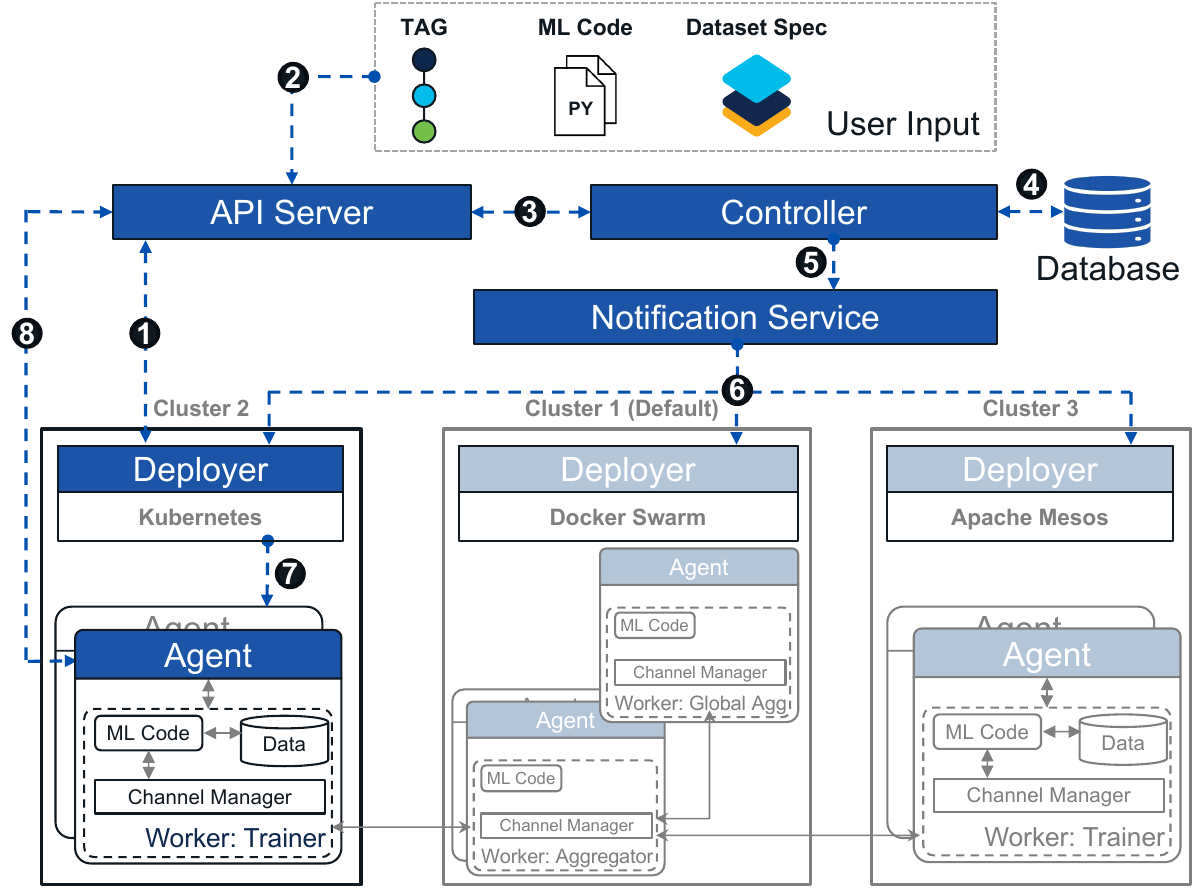}
    \caption{\sys architecture and workflow overview.}
    \label{fig:sys_workflow}
\end{figure}

\subsection{Workflow}
\label{s:workflow}
In \fig{fig:sys_workflow} we describe how \sys is used to register the available compute infrastructure and datasets and to deploy a distributed ML job across those resources. 

\noindent{\bf Compute Registration.} In order to register a compute cluster, the cluster admin is required to support the \sys's \textit{Deployer} interface in their cluster. By default, \sys already integrates the Deployer interface for Kubernetes (K8s). During the bootstrap of the deployer, the admin also assigns a name and provides properties associated with the cluster. Once the deployer is up, it uses a REST API call to register the cluster with \sys \hollowcircled{1}. This is one time process and the admin has full control of the resources provided by the cluster that can be used by \sys, which implies that the admin can deregister their cluster if needed.

\noindent{\bf Job Configuration.} To submit an FL job to \sys, the user needs to provide a job configuration that consists of three main components: (i) a TAG-based high-level abstract description of the machine learning job, (ii) program logic associated with each role, and (iii) data specification configuration containing metadata information about the datasets, which provides deployment constraints. The job is submitted to the system through APIServer \hollowcircled{2}.

\noindent{\bf Job Deployment.} Upon receiving the job configuration, it is shared with the controller \hollowcircled{3}. The controller records this information in the database \hollowcircled{4}, and expands the TAG to determine where each role should be created, based on the metadata: \textit{datasetGroups}, \textit{groupBy} and \textit{groupAssociation} attributes. The controller then sends a compute creation event to the notifier along with the job information \hollowcircled{5}. The notification service notifies the corresponding deployers where roles need to be created \hollowcircled{6}. Upon receiving such a request, each deployer creates a compute (e.g., a pod in K8s) that contains an agent \hollowcircled{7}. The agent uses the job id to retrieve the code and task configuration files \hollowcircled{8}. It then starts a worker which executes an FL task. Once the task is completed, the agent updates its status via the APIServer. The deployers are subsequently notified through a revoke deploy event to de-allocate the resources from the compute clusters. Our system manages FL jobs in a fully automated manner.

\subsection{Implementation}
\label{s:fiab}
We have implemented the management plane of \sys with 12K LOC in Golang while the \sys SDK was developed with 7K LOC in Python. The current implementation supports a diverse range of topologies as shown in \tref{tbl:flframework_comparison}. \sys also provides an emulation platform called \textbf{\sys-In-A-Box} (fiab). It is a single machine management plane that leverages minikube, a local Kubernetes cluster. All of the system components are deployed as pods on minikube. This single box deployment of our system allows users to easily validate their prototypes of new FL mechanisms and algorithms or to conduct small scale experiments. Moreover, packaging the system components as pods makes the management plane deployment portable, enabling it to be easily deployed in a real world Kubernetes cluster.

\section{Evaluation}
\label{sec:evaluation}
To demonstrate the flexibility and extensibility of \sys in supporting various FL topologies, our evaluation aims to achieve the following objectives. 

\begin{tightitemize}
	\item \secref{s:extend_topo} demonstrates how developers can extend a sample H-FL topology  (as shown in \fig{fig:tag_expand1}) for complex settings by incorporating a coordinator. 
	\item \secref{s:flexible_backend} highlights the advantages of selecting communication backends, emphasizing the system's flexibility. 
	\item \secref{subsec:topo_transformation} illustrates the ease of transforming an FL job between different topologies.
	\item \secref{s:tag_overhead} presents micro-benchmarking
          results that quantify the overhead of TAG expansion. 
    \item \secref{subsec:sys_comparision} provides quantitative and qualitative
              comparison of \sys and existing systems.
\end{tightitemize}
 
\begin{figure}[t!]
  \centering
  \includegraphics[width=0.96\columnwidth]{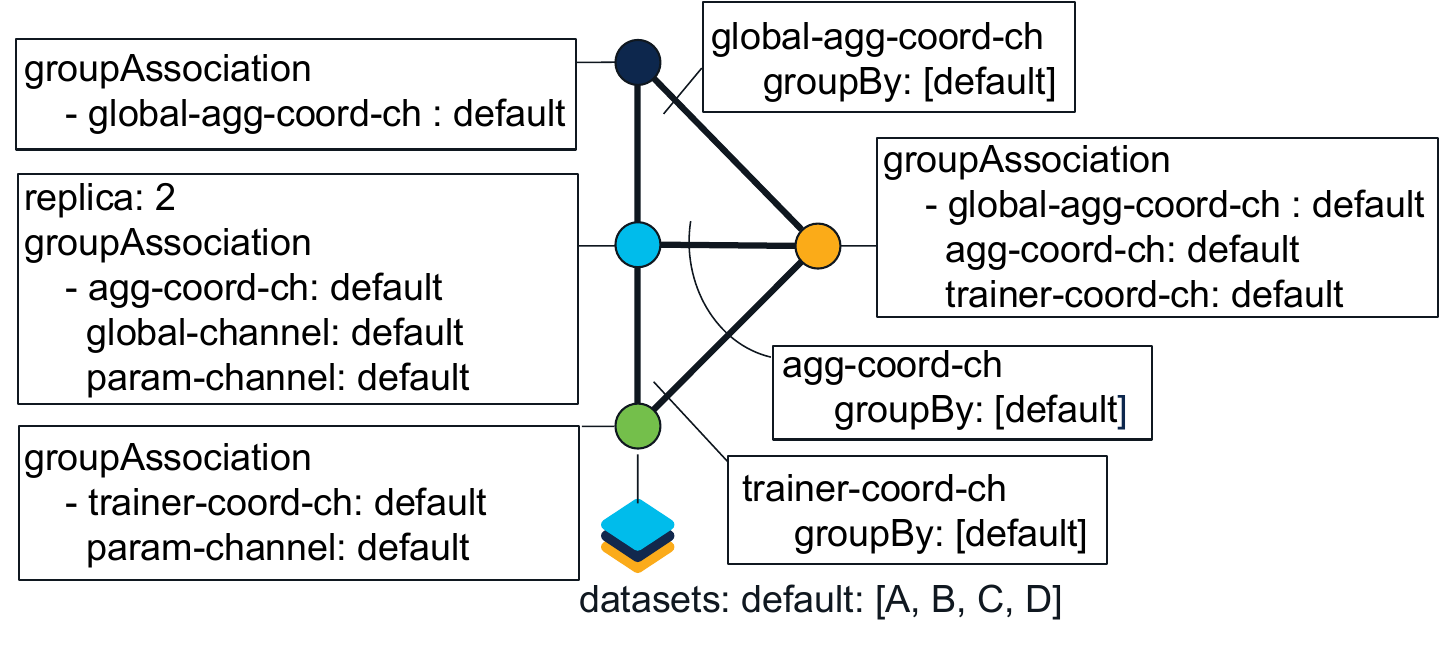}
  \caption{TAG for Coordinated FL (H-FL with coordinator). Only
    additional changes are shown in the figure on top of the
    configuration shown in \fig{fig:tag_expand1}. TAG is expressed
    in the YAML format. \mydarkbluedot: global aggregator, \mypalebluedot:
    aggregator, \mygreendot: trainer, and \myorangedot:
    coordinator. The expanded form is shown in
    \fig{fig:topo_hfl_coord}.}
  \label{fig:tag_coord_fl}
\end{figure}

\subsection{Extension for New Mechanisms}
\label{s:extend_topo}

\begin{figure}[t]
\begin{lstlisting}[frame=tb,style=yaml]
name: A CO-FL TAG example
roles:
- name: coordinator
  groupAssociation:
  - global-agg-coord-ch: default
    agg-coord-ch: default
    trainer-coord-ch: default
...
- name: aggregator
  replica: 2
  groupAssociation:
  - agg-coord-ch: default
    param-channel: default
    global-channel: default
- name: trainer
  isDataConsumer: true
  groupAssociation:
  - trainer-coord-ch: default
    param-channel: default
channels:
- name: global-agg-coord-ch
  pair: [global-aggregator, coordinator]
  groupBy:
    type: tag
    value: [default]
  funcTags:
    global-aggregator: [coordinate]
    coordinator: [coordinateWithTopAgg]
...
- name: param-channel
  groupBy:
    type: tag
    value: [default]
  pair: [aggregator, trainer]
  funcTags:
    aggregator: [distribute, aggregate]
    trainer: [fetch, upload]
\end{lstlisting}
\caption{TAG representation for CO-FL topology shown in \fig{fig:tag_coord_fl} in the YAML format. Some blocks of the configuration are omitted for brevity.}
\label{fig:example_cofl_schema}
\end{figure}

The \textit{developer programming model} and TAG mechanism of \sys facilitate the extension of topologies and the addition of new mechanisms. An example of topology extension is illustrated in \fig{fig:topo_hfl_coord}, which shows an H-FL topology with a coordinator. In this paper, we refer to this variant as Coordinated Federated Learning (CO-FL). CO-FL differs from  H-FL in two key aspects: (1) the links between aggregator and trainer form a bipartite graph in CO-FL, and (2) the coordinator is connected to the rest of the roles. Enabling this new variant requires three types of modifications: (i) an update to the TAG, (ii) an update to the implementation of roles in the H-FL TAG to allow communication with the coordinator, and (iii) dataset group update in the dataset specification.

\noindent{\bf TAG Changes.} In \fig{fig:tag_coord_fl}, we illustrate the modifications required to integrate a coordinator into the H-FL topology depicted in \fig{fig:tag_expand1}.The corresponding TAG representation is illustrated in \fig{fig:example_cofl_schema}. The transformation process entails modifying only 46 lines of configuration. The majority of the changes (36 lines, 78\%) involve configuring new channels for the coordinator, while the rest of the changes are made to existing roles and channels. The addition of a coordinator requires configuring the \textit{replica} attribute to match the number of aggregators (\secref{subsection:tag}), and allows for the creation of bipartite-like communication links upon TAG expansion. 

\begin{figure}[t]
\begin{lstlisting}[language=Python, frame=tb, numbers=left, xleftmargin=2em, framexleftmargin=2.5em]
def compose(self) -> None:
  super().compose()

  with CloneComposer(self.composer) as composer:
    self.composer = composer
    tl_coord_ends = Tasklet("get_coord_ends", self.get_coord_ends)

  tl = self.composer.get_tasklet("distribute")
  tl.insert_before(tl_coord_ends)
\end{lstlisting}
\caption{Code snippet for global aggregator for CO-FL.}
\label{fig:cofl_global_agg_code}
\end{figure}

\noindent{\bf Code Changes.} Following the completion of TAG, the next step is to implement each role in the TAG. \sys's developer programming model allows easy extension without the need for modifying the core library. The developer inherits the base classes of H-FL and implements additional functionality for the coordinator role. In CO-FL, while the global aggregator performs the same steps as it does in H-FL, it must receive information from the coordinator about which intermediate aggregators to send and receive model weights as not all intermediate aggregators may be involved in the training for every round.

  Such a functionality, implemented as a \texttt{get\_coord\_ends} tasklet, is updated into an inherited \texttt{tasklet} chain of the global aggregator. As shown in \fig{fig:cofl_global_agg_code} (lines 8 and 9), the tasklet is introduced before the \texttt{distribute} call using the API in \tref{tbl:prog_api}. We obtain tasklets by using their alias and call appropriate operations (e.g., \texttt{insert\_before}, \texttt{remove}). Other roles are implemented in a similar fashion, resulting in minor code revisions for the CO-FL implementation, as shown in \tref{tbl:sys_loc_comparision}.

\begin{figure}[t]
\begin{lrbox}{\mylistingbox}%
\begin{minipage}{0.46\columnwidth}%
\begin{lstlisting}[style=yaml]
name: Dataset A
url: someurl/data_A.npz
realm: cluster1
isPublic: true
\end{lstlisting}%
\end{minipage}%
\end{lrbox}%

\subfloat[An example dataset]{\label{fig:example_cofl_dataset_reg}\usebox{\mylistingbox}}%
\begin{lrbox}{\mylistingbox}%
\begin{minipage}{0.54\columnwidth}%
\begin{lstlisting}[style=yaml,xleftmargin=10pt]
- role: trainer
  datasetGroups:
    default: [A, B, C, D]

\end{lstlisting}%
\end{minipage}%
\end{lrbox}%
\subfloat[Dataset specification]{\label{fig:example_cofl_dataspec}\usebox{\mylistingbox}}%
\caption{Metadata information used in mapping dataset to a correct role and dataset group.}
\label{fig:example_cofl_dataset_dataspec}
\end{figure}

\noindent{\bf Dataset Group Update.} Data owners can register their datasets by providing metadata information. The dataset registration process is independent of composing a job. An example for the metadata is shown in \fig{fig:example_cofl_dataset_reg}. Upon registration, \sys produces an ID for the metadata. Assuming that datasets are already registered in the system with their IDs as $A$, $B$, $C$ and $D$, ML practitioners can use the IDs and group the datasets to different dataset groups. In the H-FL example (\fig{fig:tag_expand1}), $A$ and $B$ belonged to the $west$ group and $C$ and $D$ belonged to the $east$ group. The modification to support the example of \fig{fig:tag_coord_fl} is to group the four datasets into the default group for the trainer role (\fig{fig:example_cofl_dataspec}).

\begin{figure}[t]
  \centering
  \includegraphics[width=0.40\textwidth]{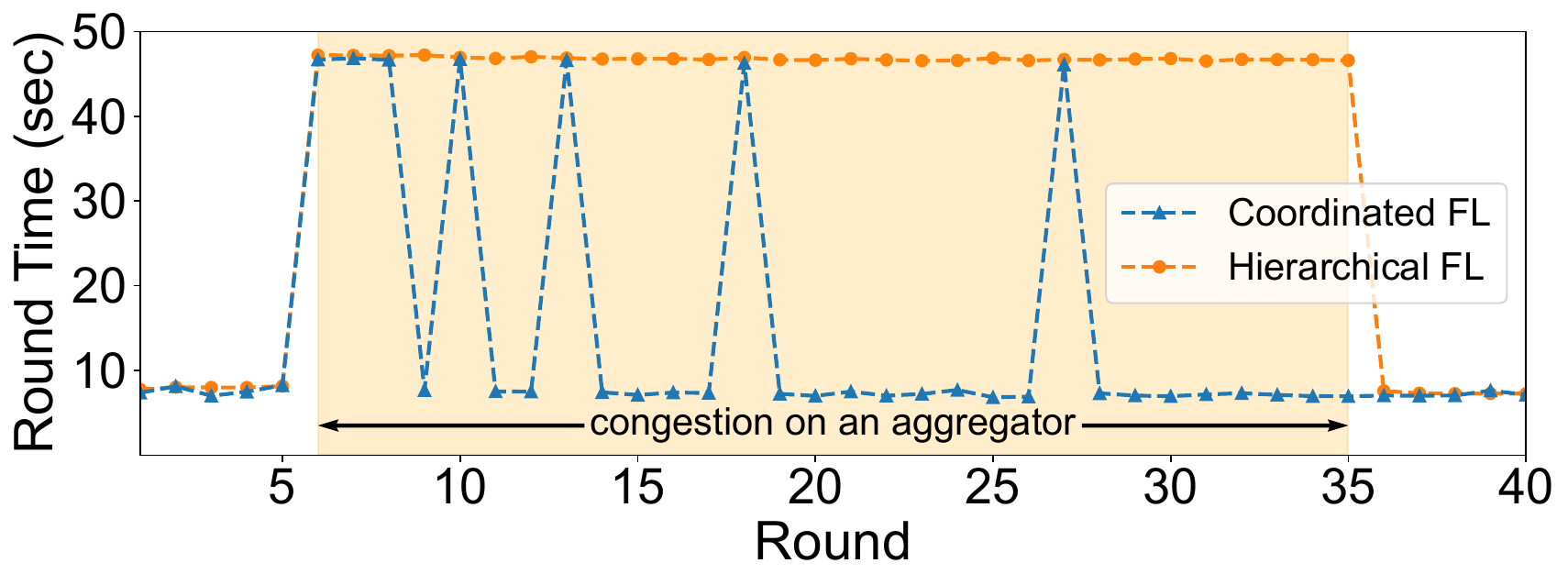}
  \caption{Performance comparison between Coordinated FL vs Hierarchical 
  	FL. Coordinated FL manages the network congestion with its load-balancing
  	scheme.}
  \label{fig:coordexp}
\end{figure}

\noindent{\bf Setup.} 
To illustrate the feasibility of CO-FL extension, we implement a toy scenario with 10 trainers, two aggregators, and a coordinator, where a link between the global aggregator and an aggregator becomes a bottleneck over multiple rounds. The coordinator identifies and exclude a slow aggregator as the delay is reported.

\noindent{\bf Result.} 
\fig{fig:coordexp} compares the results of such a scenario with H-FL. In CO-FL, from round \#6, the coordinator detects a delayed aggregator based on model upload times. After three consecutive rounds of delays, the slow aggregator is disabled for one round (\#9), checked for delay in round \#10, and, if delayed, disabled for two rounds (\#11-12). As the delay persists, the coordinator disables the straggler in a binary-backoff manner. Without the coordinator, H-FL suffers extended per-round times due to the straggler.

\begin{figure}[t]
 \centering
 \subfloat[MNIST dataset]{\label{fig:hybridexp_mnist}\includegraphics[height=1in]{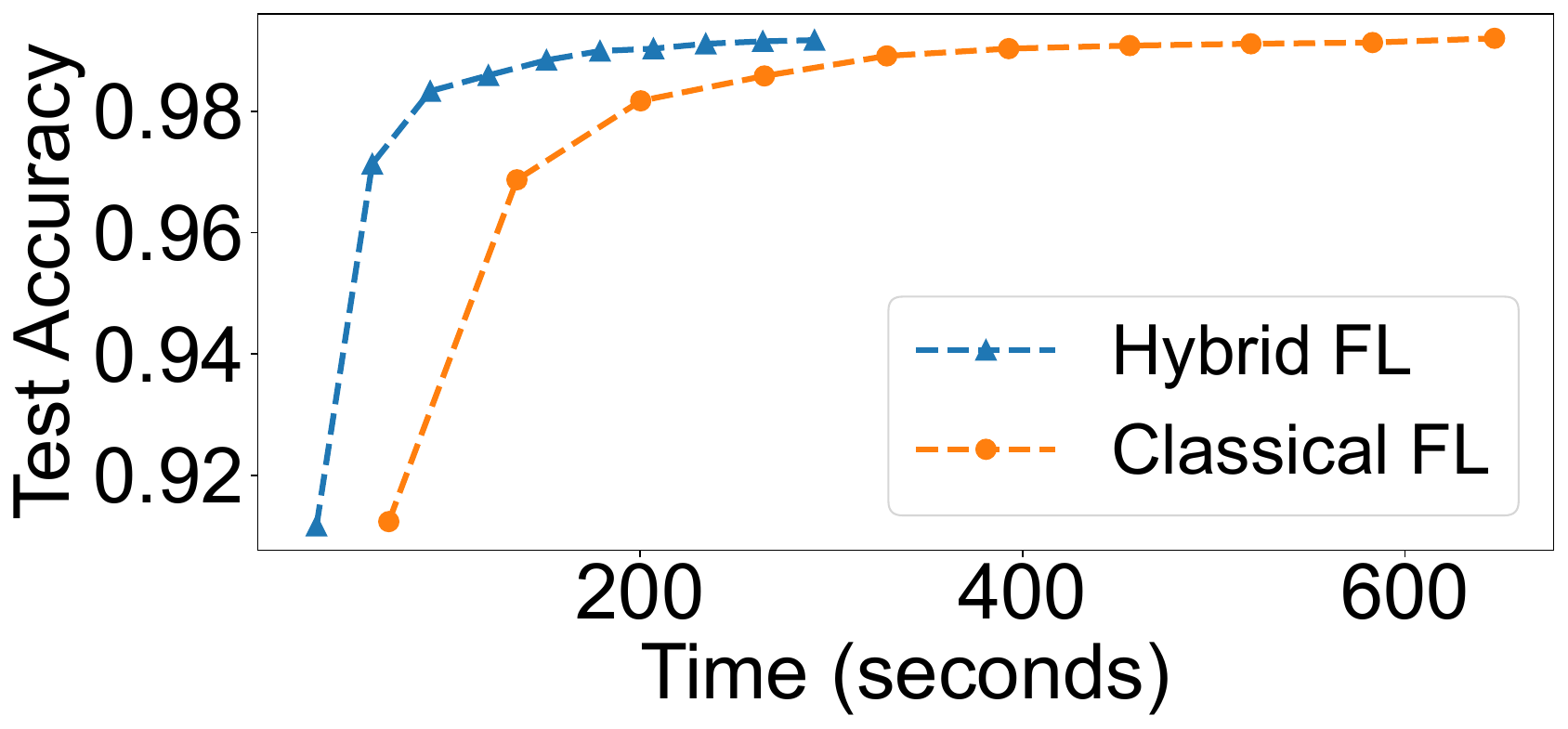}}\hfill
 \subfloat[CIFAR-10 dataset]{\label{fig:hybridexp_cifar10}\includegraphics[height=1in]{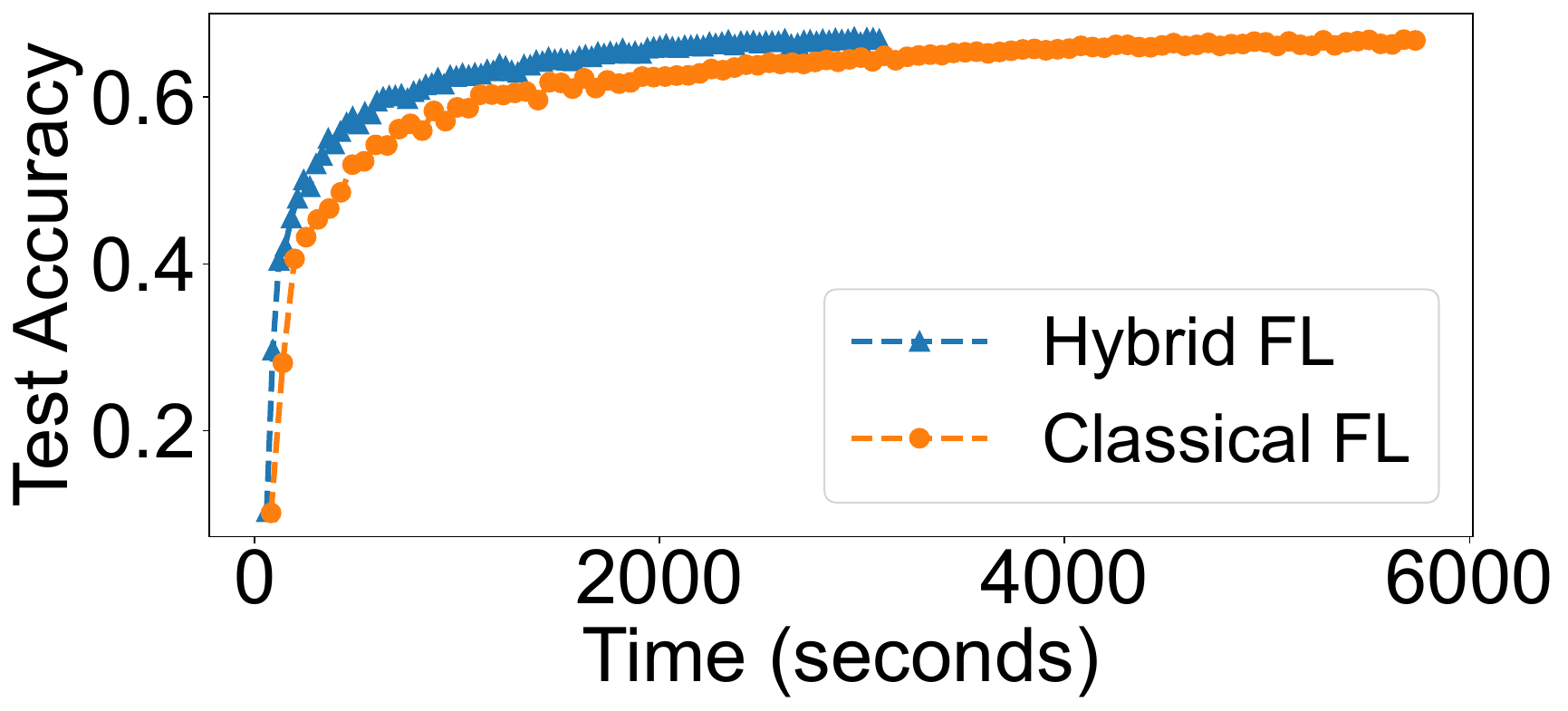}}
 \caption{Performance comparison between Hybrid FL vs Classical
   FL. \sys's flexibility in communication backend selection demonstrates the
   efficacy of Hybrid FL.}
 \label{fig:hybridexp}
\end{figure}

\subsection{Flexible Backend}
\label{s:flexible_backend}

We demonstrate the versatility of \sys's backend configurations and their implications by implementing hybrid FL~\cite{hybrid_fl} (\fig{fig:topo_fl_hybrid}). Hybrid FL is good for scenarios where trainers are co-located and network bandwidth among trainers are much higher than bandwidth between trainers and the aggregator. In hybrid FL, instead of individual trainers sending their model updates, co-located trainers form clusters and share their model updates with the aggregator. Within each cluster, trainers utilize the ring-allreduce algorithm~\cite{ring_allreduce} to exchange their model updates, resulting in a single copy of the cluster-level model being shared with the aggregator.

\noindent{\bf Setup.} In contrast to frameworks limited to a single backend configuration, \sys offers the flexibility to configure the TAG with multiple distinct backends. The hybrid FL example uses a gRPC backend (P2P) for transferring model updates between trainers within the cluster, and an MQTT backend for communication between the aggregator and different clusters. We create a hybrid topology that consists of 50 trainers and emulate different bandwidth on each backend, by utilizing the Linux \texttt{tc} tool. We used the MNIST~\cite{mnist} and CIFAR-10~\cite{cifar10} datasets for the experiments, training a 2-layer CNN model and a ResNet-18 model respectively. The trainers are equally divided into five groups and a trainer within one of the cluster is chosen as a straggler where its bandwidth is configured to 1~Mbps for the MNIST and 10~Mbps for the CIFAR-10 experiment. While all the other gRPC backend is given a maximum bandwidth of 100~Mbps. Additionally, for comparison, we set up a C-FL topology using MQTT as the backend with 50 trainer nodes, where one trainer is designated as a straggler.

\noindent{\bf Result.} \fig{fig:hybridexp} shows the test accuracy over wall-clock time, where each point represents a round. The results suggest that hybrid FL converges faster than C-FL, by achieving 2.21$\times$ and 2.01$\times$ speedup in reaching 0.985 and 0.650 accuracy in MNIST and CIFAR-10 datasets. This is primarily because hybrid FL allows non-straggling trainers to efficiently upload cluster-aggregated weights to the global aggregator. In contrast, in the case of C-FL, the whole system must wait for the straggler's weight upload to the global aggregator. Hybrid FL also consumes less bandwidth compared to C-FL to upload model updates (25~MB vs 250~MB/round for MNIST and 223~MB vs 2230~MB/round for CIFAR-10). This experiment shows that \sys allows flexible communication backend configurations and such flexibility can help design and evaluate new FL approaches easily and rapidly.


\begin{table*}[t!]
  \small
  \centering
  \ra{1.2}
 \scalebox{0.88}{
  \begin{tabular}{c c c c c c c}
    \toprule
    & {C-FL} & {C-FL$\rightarrow$H-FL} & {H-FL$\rightarrow$H-FL$^b$} & {C-FL$\rightarrow$Distributed} & {C-FL$\rightarrow$Hybrid} & H-FL$\rightarrow$CO-FL \\
    \midrule
    Code & \makecell{+ trainer \\ + global-agg} & + agg (25$^\dag$) & N/A & \makecell{- global-agg \\ $\Delta$ inheritance (1)} & \makecell{ $\Delta$ inheritance (2)} & \makecell{+ coordinator (13$^\dag$) \\ $\Delta$ inheritance (3)}
    \\
    \midrule
    TAG  & + channel & \makecell{+ channel (16+$N$) \\ + role (3+2$N$) \\ $\Delta$ groupAssociation ($N$)} & \makecell{$\Delta$ groupBy ($N$) \\ $\Delta$ groupAssociation ($N$)} & $\Delta$ channel (2) & \makecell{+ channel (12+$N$) \\ $\Delta$ groupAssociation (2$N$)} & \makecell{+ replica (1) \\ + role (5) \\ + channels (42) \\ $\Delta$ groupAssociation (3)}
    \\
    \midrule
    Metadata & + init info & $\Delta$ datasetGroups ($N$) & $\Delta$ datasetGroups ($N$) & N/A & $\Delta$ datasetGroups ($N$) & $\Delta$ datasetGroups (1)
    \\
    \bottomrule
  \end{tabular}
  }
  \caption{Changes required to transform from one topology to another. Given TAG in the YAML format, required lines of code changes are shown in parentheses. $N$ denotes the number of groups. The TAG representation of C-FL, H-FL, Distributed and Hybrid is showcased in \fig{fig:topo_tag_examples} while TAG for CO-FL (\fig{fig:topo_hfl_coord}) is shown in \fig{fig:tag_coord_fl}. H-FL$^b$ represents H-FL topology with different grouping options. +, - and $\Delta$ represent addition, removal and update respectively and N/A indicates no change. $^\dag$: only needs template code.
  ``$\Delta$ inheritance'' implies the switch of a base class from one to another.}
  \label{tbl:topo_transformation}
\end{table*}

\subsection{Topology Transformation: User Perspective}
\label{subsec:topo_transformation}

The requirements and constraints for the ML job may change over time, which may require changes in learning topology. To demonstrate how easily these transformations can be made, we walk through the steps of transforming from one topology to another, starting with a basic C-FL topology. Note that these transformations are done offline and we leave the dynamic reconfiguration of topology for future investigation. The transformation steps presented here only involve user application code and TAG changes. In contrast, \secref{s:extend_topo} discussed how to extend the logic for new topology.

\noindent{\textbf{Classical$\rightarrow$Hierarchical.}} C-FL topology consists of trainers and global aggregator. To transform from C-FL to H-FL, a user needs to introduce an (+) aggregator role, a new connecting (+) channel with the new aggregator. Finally, to allow the grouping of trainer nodes the ($\Delta$) \textit{datasetGroups} attribute in metadata information is updated. These modifications each require only a change of up to 16 lines of code (LOC), with additional updates that varies on the number of newly introduced aggregators.

\noindent{\textbf{Classical$\rightarrow$Distributed.}} In FL, trainers send their model weights to the aggregator while in distributed learning they are shared among all the nodes directly. \sys SDK provides a separate trainer base class for federated and distributed learning. Thus, from the user's perspective, C-FL to distributed training change requires, removing the global aggregator, ($\Delta$) updating the inherited base trainer class (1 LOC), and ($\Delta$) altering TAG representation (2 LOC) where the trainer-aggregator channel is updated to trainer-trainer channel as shown in \fig{fig:topo_tag_dist}. 

\begin{table}[t]
  \centering
	\scalebox{0.73}{
  \begin{tabular}{cccccccc}
    \toprule
    \multirow{2}{*}{Topology}&\multicolumn{1}{c}{\multirow{2}{*}{Task}}&\multicolumn{6}{c}{Number of Workers}\\
    \cmidrule{3-8}
    &&\multicolumn{1}{c}{1}&{10}&{100}&{1,000}&{10,000}&{100,000}\\
    \midrule
    \multirow{2}{*}{Classical FL}&Expansion&0.005&0.006&0.036&0.329&3.183&31.990
    \\
    &DB Write&0.007&0.008&0.037&0.315&2.781&27.971
    \\
    \midrule
    \multirow{2}{*}{Coordinated FL}&Expansion&0.006&0.012&0.041&0.320&3.190&32.538
    \\
    &DB Write&0.033&0.035&0.061&0.317&2.901&27.232
    \\
    \bottomrule
  \end{tabular}
  }
  \caption{TAG expansion latency in seconds.}
  \label{tbl:expansion_overhead}
\end{table}

\noindent{\textbf{Classical$\rightarrow$Hybrid.}}
Transformation from C-FL to hybrid topology entails two steps: First, it would require ($\Delta$) updating the inherited trainer and global aggregator class. Again, the \sys SDK provides base classes for hybrid topology, thus, a user just needs to change the inherited parent class name in the trainer and global aggregator role's program (1 LOC each). Then, it needs ($\Delta$) to change the TAG to create appropriate channels with backends and change \textit{groupBy} and \textit{datasetGroups} to group co-located datasets, which only requires up to 12 LOC change with additional updates on varying number of trainer groups.

\noindent{\textbf{Hierarchical$\rightarrow$Coordinated.}}
CO-FL is different from H-FL in that a coordinator oversees a federated learning process. Therefore, in CO-FL, a user needs to introduce the coordinator (13 LOC), ($\Delta$) update the inheritance of classes for the global aggregator, aggregator, and trainer (1 LOC each), add coordinator role (5 LOC), and add new channels between the coordinator and the rest of the roles (14 LOC each). In addition, grouping between aggregators and trainers can be dynamic based on the coordinator's logic. For that, the user ($\Delta$) updates \textit{datasetGroups} as a single group (1 LOC) and configures \textit{replica} in the TAG (1 LOC). Note that the reported LOC is for user application code, not the logic for CO-FL.

\subsection{TAG Expansion Overhead}
\label{s:tag_overhead}
TAG expansion is the first step in the management plane for deployment preparation. Deployment time varies based on factors like network bandwidth, cluster resources, and job size; our evaluation focuses on TAG expansion overhead rather than deployment issues in geo-distributed scenarios. We conducted experiments to measure the latency of TAG expansion and database write of its results on \sys for two FL topologies (C-FL and CO-FL) by varying the number of trainers. CO-FL was configured with 100 replicas and a coordinator. The results shown in \tref{tbl:expansion_overhead} demonstrate that the overhead of TAG expansion on \sys is comparable across the two FL topologies. The results also show that \sys is highly scalable, achieving TAG expansion on 100,000 trainers under a minute for both FL topologies. The current implementation can be further optimized since it only uses a single CPU core and data is duplicated during the expansion.


\subsection{System Comparison}
\label{subsec:sys_comparision}

\subsubsection{Comparison with FedML}
\label{s:comp_fedml}

FedML uses client-server architecture which limits the system's ability to be extensible and support  flexible topologies. FedML provides native support for C-FL and H-FL ($n=1$) and wraps the underlying implementation of model, data loading and component logic as part of the core library. Unlike \sys, that allows a user to define their own model and data loader through the  \textit{programing model}; users in FedML provide the model and data information through configuration files and introduction of new data loader/model or component requires changes in the core library. Note that, to support H-FL, FedML chooses to modify the training \verb|client manager| class with the appropriate functionality, and a hardcoded method to allow a worker to distriguish whether it needs to act as a trainer or a middle aggregator. This is philosophically different from \sys where we implement a new middle aggregator role and keep the global aggregator and trainer untouched (see LOC changes for H-FL in \tref{tbl:sys_loc_comparision}). 

To further compare the flexibility provided by FedML and \sys, we leverage the native C-FL and H-FL implementations in FedML and extend them to support two new topology (i) $n$-level H-FL where $n$ represents the number of intermediate aggregators, and (ii) CO-FL as described in \secref{s:extend_topo}. \tref{tbl:sys_loc_comparision} illustrates the effort in terms of LOC required to implement different topologies. 

\begin{table}[t]
\centering
  \scalebox{0.70}{
  \begin{tabular}{l c c c c c>{\raggedright\arraybackslash}p{1.62cm}}
  \toprule
	Topology & System & Global Agg. & Agg. & Trainer & Coordinator & Core Lib Changes \\
	\midrule
	\multirow{2}{*}{C-FL*} & \sys & 231 & --- & 156  & ---  & \multicolumn{1}{c}{\xmark} \\
                  & FedML  & 577 & --- & 319 & --- & \multicolumn{1}{c}{\xmark}  \\
	\midrule

	\multirow{2}{*}{H-FL*} & \sys & 0 & 200 &  0 & ---  & \multicolumn{1}{c}{\xmark} \\
                  & FedML  & 0 & 0 & 68 & --- & \multicolumn{1}{c}{\xmark} \\
	\midrule

	\multirow{2}{*}{\shortstack{N-level\\H-FL}} & \sys & 0 & 0 & 0  & ---  & \multicolumn{1}{c}{\xmark} \\
                  & FedML  & 0 & 190 & 0 & --- & \multicolumn{1}{c}{\cmark} \\
	\midrule

	\multirow{2}{*}{CO-FL} & \sys & 40 & 67 & 73  & 158  & \multicolumn{1}{c}{\xmark}  \\
                  & FedML  & \multicolumn{4}{c}{Not Achieved (NA)} & \multicolumn{1}{c}{\cmark} \\
	\bottomrule
\end{tabular}
 }
\caption{Comparing development effort in number of lines of code between \sys and FedML to enable new topologies by adding/modifying roles. H-FL extends C-FL while $n$-level H-FL and CO-FL extends H-FL. *: natively supported.}
  \label{tbl:sys_loc_comparision}
\end{table}

\noindent{\bf{$n$-level H-FL.}} Extending the base H-FL to add new intermediate aggregators in \sys is trivial. It requires changes only in the TAG to introduce new aggregator roles and $0$ LOC changes for role's code. FedML on the contrary, requires significant code changes 190 LOC, across multiple files which are part of the core-library. These changes are required to overcome three limitations. First, the native \verb|client manager| can only synchronize model weights outside its silo (group) with the global aggregator. In $n$-level H-FL, the $n-1$ level intermediate aggregators synchronize weights with aggregators in different silos, while only the top-most middle aggregator level synchronizes weights with the global aggregator. Secondly, hardcoding in \verb|client manager| for communication with the global aggregator needed to be removed, as the native H-FL ($n=1$) implementation enforced all middle aggregators to share updates to the global aggregator alone. Thirdly, while the native implementation provided means to organize trainers in data silos, we needed to implement a new configuration mechanism to allow grouping of different aggregators in the $n$-level H-FL topology. 

\noindent{\bf{CO-FL.}} Implementing CO-FL requires changes in the H-FL TAG configuration and code updates associated with different roles as described earlier and shown in \tref{tbl:sys_loc_comparision}. However, implementing such a topology using FedML requires more fundamental implementation changes. In case of CO-FL the coordinator assigns a trainer to intermediate aggregator on the fly. However, FedML uses a torch~\cite{pytorch} backend and its process-groups for model update synchronization, and this makes it difficult for FedML to be easily extensible to CO-FL. This is because process-groups are created per silo, and weights are broadcast within the group. Thus, separating out trainers into different groups on the fly is not  trivial. It requires significant development effort to support CO-FL in FedML and is marked as not achieved (NA) in \tref{tbl:sys_loc_comparision}.


\begin{figure}[t]
 \centering
 \subfloat[Accuracy]{\label{fig:flame_fedml_flower_round_to_accuracy}\includegraphics[height=1in]{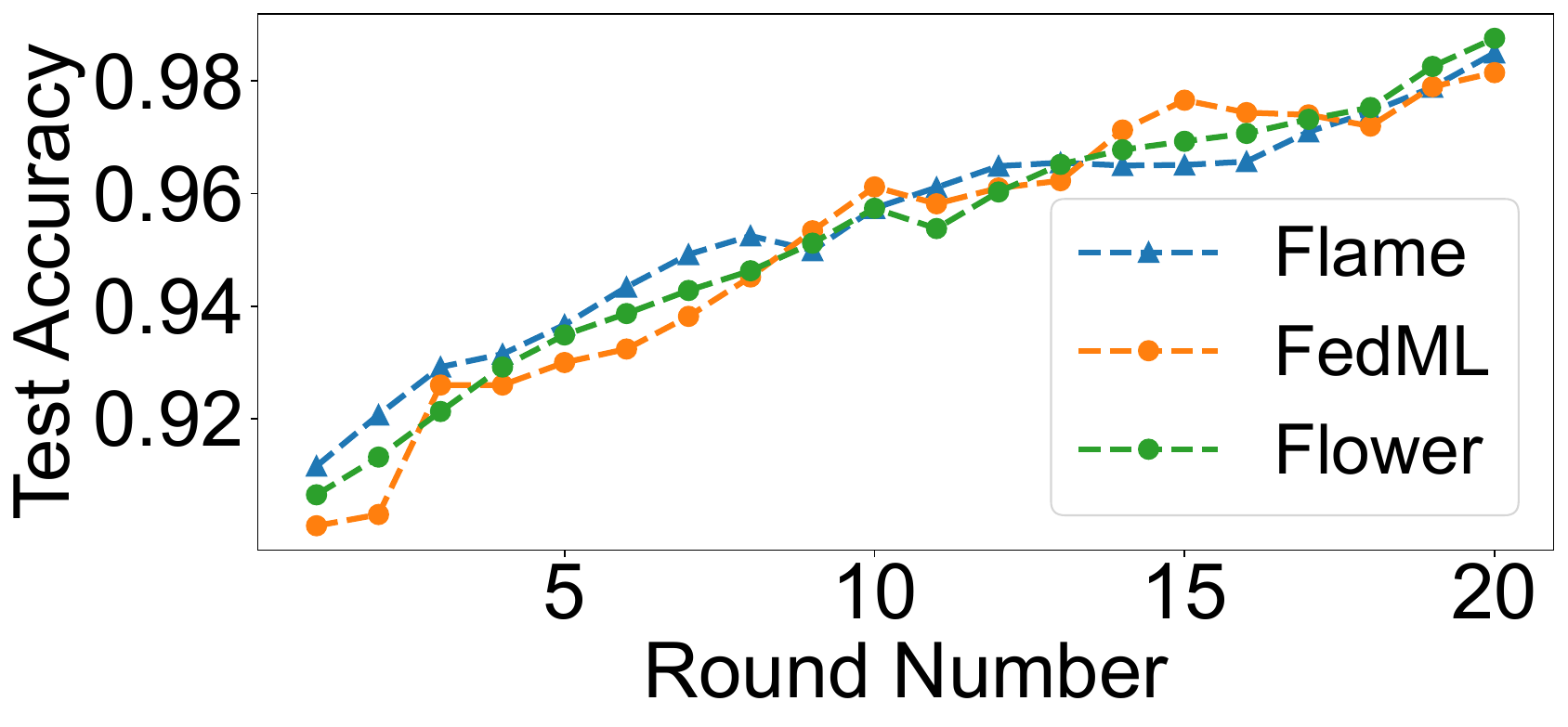}}\hfill
 \subfloat[Time per round]{\label{fig:flame_fedml_flower_per_round_time}\includegraphics[height=1in]{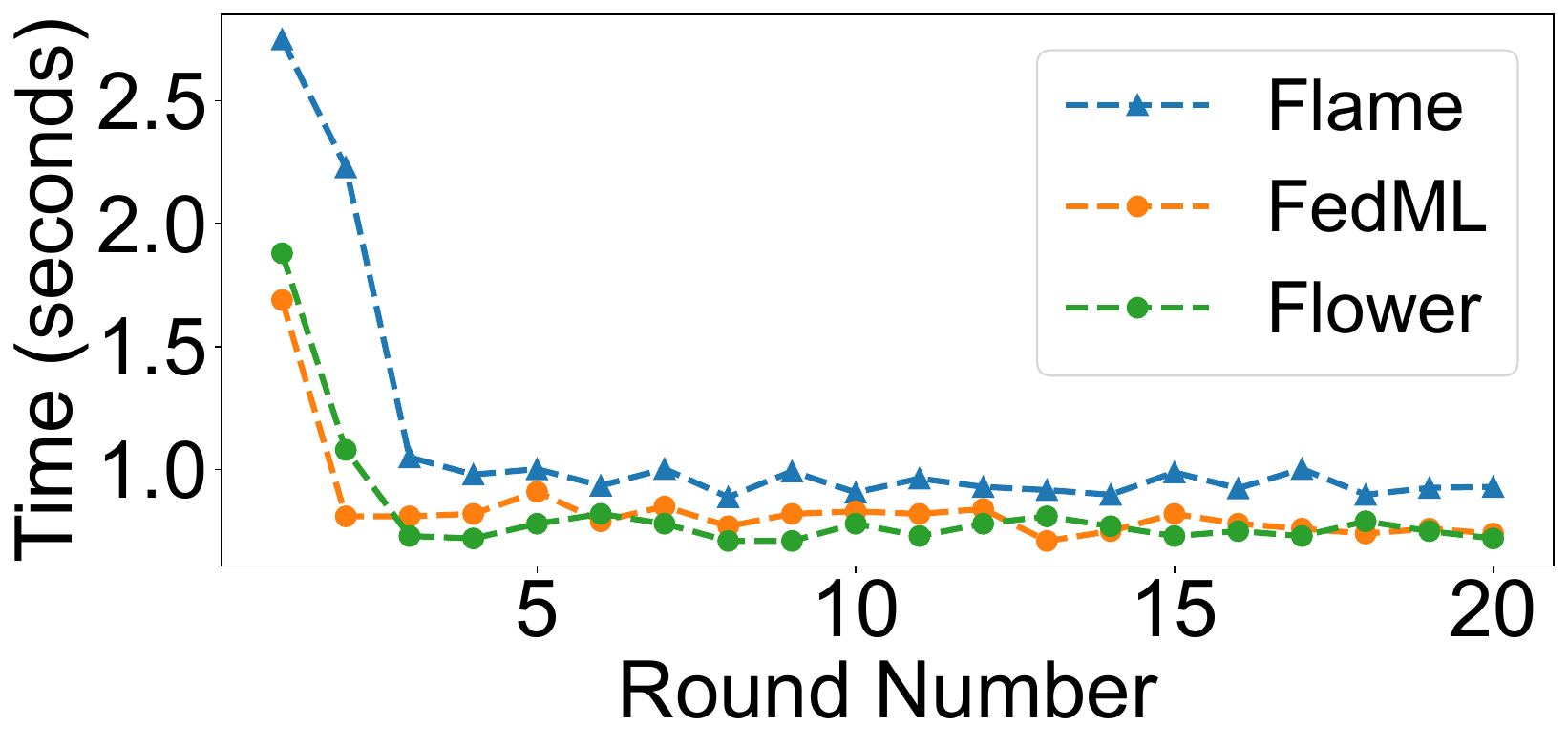}}
 \caption{Performance comparison with FedML and Flower.}
 \label{fig:flame_fedml_flower_perf_comparision}
\end{figure}

\noindent{\bf{Performance.}} Finally, we compare \sys with FedML in terms of accuracy and per-round execution time under the C-FL topology with the MNIST dataset. We also add Flower's results as another reference point. \fig{fig:flame_fedml_flower_perf_comparision} demonstrates that all of them achieve comparable performance. 

\subsubsection{General Comparison}
We compare other FL frameworks by considering their native support for different topologies and support for communication backend between components. For brevity, we focus on highlighting the key differences among FedML, Flower, and FedScale, as presented in \tref{tbl:sys_loc_comparision}, while omitting other differences with respect to training and aggregation algorithms, client selection algorithms, or supported monitoring capabilities. 

Out of the nine listed topologies, \sys natively supports eight of them, whereas the other frameworks only support a few of them. Although \sys does not natively support vertical FL topology, it can be easily implemented due to its extensibility, as demonstrated through the CO-FL use case. However, extending FedML to accommodate additional topologies presents challenges, as discussed earlier. Similar additional effort is required to build other topologies in the other frameworks.

Another crucial distinction between \sys and the other frameworks is that \sys supports different communication protocols on a \textit{per-channel} basis. Both \sys and FedML facilitate integration of different communication protocols, such as MQTT and gRPC, while the other frameworks only support gRPC. The logical graph abstraction employed by \sys breaks down the connections between different roles (workers) into channels, enabling per-channel communication control for any FL topology. In contrast, the other frameworks enforce the use of the same communication backend for all connections between nodes within a job. 

Finally, support for deploying an FL job is classified into two approaches --- compute centric v/s compute agnostic. In \sys, the deployer component enables the system to connect with different compute clusters managed by various resource orchestrators. This capability allows for a \textit{compute-agnostic} approach, where the user provides the ML code and deployment instructions/rules, and the system takes care of locating the appropriate compute units, creating the group, deploying the code, and initiating the learning process. In contrast, other frameworks follow a \textit{compute centric} approach, which requires the participant to select appropriate compute resources based on constraints associated with the data, such as GDPR rules, and to manually create groups in the case of H-FL, thereby making them less extensible than \sys.

\begin{table}[t]
\centering
  \scalebox{0.75}{
  \begin{tabular}{ccccc}
    \toprule
    \textbf{Feature} & \textbf{\sys} & \textbf{FedML~\cite{fedml}} & \textbf{Flower~\cite{flower}} & \textbf{FedScale~\cite{fedscale}}
    \\
    \midrule
    C-FL~\cite{fedavg} & \cmark & \cmark & \cmark & \cmark
    \\
    H-FL~\cite{hierFAVG_topology} & \cmark & \cmark & \xmark & \xmark
    \\
    N-level H-FL~\cite{demystifying} & \cmark & \xmark & \xmark & \xmark
    \\
    Hybrid~\cite{hybrid_fl} & \cmark$^*$ & \xmark & \xmark & \xmark
    \\
    CO-FL~\cite{fl_at_scale} & \cmark$^*$ & \xmark & \xmark & \xmark
    \\
    Vertical~\cite{vfl} & \xmark & \cmark & \xmark & \xmark
    \\
    Async H-FL & \cmark & \xmark & \xmark & \xmark
    \\
    Async CO-FL & \cmark & \xmark & \xmark & \xmark
    \\
    Distributed~\cite{fedml} & \cmark & \cmark & \xmark & \xmark
    \\
    \midrule
    gRPC & \cmark & \cmark & \cmark & \cmark
    \\
    MQTT & \cmark & \cmark & \xmark & \xmark
    \\
    MPI & \xmark & \cmark & \xmark & \xmark 
    \\
    NCCL & \xmark & \cmark & \xmark & \xmark 
    \\
    \bottomrule
  \end{tabular}
  }
  \caption{FL framework comparison in terms of topology and protocol; $^*$: a simplified version of the original design.}
  \label{tbl:flframework_comparison}
\end{table}

\section{Related Work}

\noindent{\textbf{Library.}} Machine learning libraries provide lower-level interfaces for concisely expressing models. They provide a collection of pre-built algorithms, functions, and tools for developing, training, and deploying machine learning models. TensorFlow~\cite{tensorflow}, PyTorch~\cite{pytorch}, and scikit-learn~\cite{scikit_learn} are some of the ML libraries providing lower-level interfaces for concisely expressing ML models, with the ability to create custom models and learning algorithms. These libraries are used to create ML models from the ground up while users need to build their system and integrate it with these models. \sys allows developers to use any such ML libraries. 

\vspace{0.2em}
\noindent{\textbf{Frameworks.}} Spark ML~\cite{spark} and Apache MXNet~\cite{mxnet} are open source frameworks mainly for distributed learning. Systems such as Flower~\cite{flower}, FedScale~\cite{fedscale},  and PySyft~\cite{pysyft} provide low level APIs which make them flexible. Unlike \sys, they cannot be easily extended to support different deployment scenarios as they lack suitable abstractions. OpenFL~\cite{openfl} is another FL framework based on a client-server architecture with two components: (1) collaborator, which uses a local dataset to train global models, and (2) aggregator, which receives the model updates and combines them to create the global model. Nvidia Clara~\cite{nvidia_clara} is an application framework specifically designed for healthcare use cases. There are other FL frameworks like FedML, which are based on client-server architecture and lack support for diverse FL configurations, required to express and extend the evolving deployment requirements.

\vspace{0.2em}
\noindent{\textbf {Simulators.}} Machine learning simulators enable quick testing of various machine learning algorithms, models, and techniques in a simulated environment. FedJAX~\cite{fedjax} is a research-focused federated learning simulator that provides an API for building and training machine learning models using a variety of federated learning algorithms. Flute~\cite{flute} is another federated learning simulator that focuses on scalability and efficiency. FLSim~\cite{flsim} is also a federated learning simulator that allows users to explore the effects of different federated learning algorithms and hyperparameters on model performance. \sys does not provide a simulator but it supports small scale emulation via the \sys-in-a-box.
\section{Conclusion}
We introduce \sys, a system that enables composability and extensibility for federated learning topologies. It relies on a logical TAG representation of the physical topology that exposes new capability to explicitly specialize the behavior and configuration of individual components in any learning system. Its programming model facilitates easy extensions without requiring any modification of its core library. It also provides basic support that makes it possible to deal with heterogeneous deployment environments. 

We open sourced the system to help researchers and developers build FL applications in modular fashion and to accelerate the progress of federated learning. In addition, \sys and its TAG abstraction, can be extended to support other learning architectures, including distributed or collaborative learning~\cite{colla, canoe}. Finally, the current implementation of the system assumes that topology changes are made offline, prior to deploying a job. The flexibility provided by \sys is also a step toward enabling  dynamic transitions between topologies in real time, such as in response to failures and load fluctuations, and to provide robustness via integration in CI/DI pipelines; we leave this for future work.  

\section*{Acknowledgments}

We would like to thank Gaoxiang Luo for his assistance at the early stage of this project. This research was partially supported by NSF projects CCF-2217070 and CNS-1909769, by funding from Cisco Research, and using experimental infrastructure from the NSF Chameleon Cloud testbed.


\bibliographystyle{ACM-Reference-Format}
\balance
\bibliography{reference}

\end{document}